\documentclass[letterpaper, 10 pt, journal, twoside]{IEEEtran}

\IEEEoverridecommandlockouts                              

\usepackage{graphics} 
\usepackage{epsfig} 
\usepackage{times} 
\usepackage{amsmath} 
\usepackage{amssymb}  

\usepackage[utf8]{inputenc} 
\usepackage[T1]{fontenc}    

\usepackage[dvipsnames]{xcolor}
\usepackage{hyperref}
\usepackage{url}            
\usepackage{booktabs}       
\usepackage{amsfonts}       
\usepackage{nicefrac}       
\usepackage{microtype}      

\usepackage{cite}

\usepackage{graphicx}
\usepackage{subfigure}
\usepackage{booktabs} 
\usepackage{xcolor}
\usepackage{multirow}
\usepackage{pifont}
\usepackage[noend]{algpseudocode}
\usepackage{setspace}
\usepackage{comment}
\usepackage[ruled]{algorithm2e}

\title{Learning from Imperfect Demonstrations \\ from Agents with Varying Dynamics
}

\author{Zhangjie Cao$^{*1}$ and Dorsa Sadigh$^{2}$
\thanks{Manuscript received October 15, 2020; revised January 26, 2021; accepted February 23, 2021.}
\thanks{This paper was recommended for publication by Editor Dana Kulic upon evaluation of the Associate Editor and Reviewers' comments.}
\thanks{$^{1}${\tt\small caozj@cs.stanford.edu}, and $^{2}${\tt\small dorsa@stanford.edu}. The authors are with the Department of Computer Science, Stanford University, Stanford, CA 94305, USA}
\thanks{Digital Object Identifier (DOI): see top of this page.}
 }

\begin{document}

\maketitle
\thispagestyle{empty}
\pagestyle{empty}

\begin{abstract}
Imitation learning enables robots to learn from demonstrations. Previous imitation learning algorithms usually assume access to optimal expert demonstrations.
However, in many real-world applications, this assumption is limiting. Most collected demonstrations are not optimal or are produced by an agent with slightly different dynamics. We therefore address the problem of imitation learning when the demonstrations can be sub-optimal or be drawn from agents with varying dynamics.
We develop a metric composed of a feasibility score and an optimality score to measure how useful a demonstration is for imitation learning.
The proposed score enables learning from more informative demonstrations, and disregarding the less relevant demonstrations. Our experiments on four environments in simulation and on a real robot show improved learned policies with higher expected return.
\end{abstract}

\section{Introduction}

Today's imitation learning algorithms often assume access to expert demonstrations~\cite{bain1995framework,ziebart2008maximum,ho2016generative,cao2020reinforcement}. 
However, this assumption is limiting in many real-world environments. In practice, it is expensive to obtain large number of expert demonstrations, but we usually have access to a plethora of imperfect demonstrations --- demonstrations that can range from random noise or failures to expert or even optimal demonstrations, and demonstrations that are collected from agents with different dynamics, such as different embodiments, body schema, or joint or rigid body structures.
This is even evident when collecting human demonstrations on a robot as we often have access to more suboptimal data from humans due to factors such as their bounded rationality~\cite{simon1997models} or difficulty of controlling robots with different or high degrees of freedom~\cite{losey2020controlling}.

Prior work has studied learning from suboptimal demonstrations by assigning an optimality score for each demonstration, and only relying on learning from the optimal data~\cite{pmlr-v97-wu19a,grollman2011donut,zhu2020learning,tangkaratt2019vild}.
In this paper, we go beyond this assumption and consider imperfect demonstrations that can also be collected from agents with different dynamics, e.g., demonstrations on a robot with different number of joints than our target agent. 
In this setting, simply considering optimality can fail because some demonstrations assessed as optimal might not even be feasible on the target robot. 
Our key insight is that we need to measure \emph{optimality} along with \emph{feasibility} of demonstrated trajectories to effectively learn policies for a target domain.
This further requires us to go beyond existing optimality measures, as an optimal demonstration in another dynamics may not be optimal for our target agent.

\begin{figure}[ht]
    \centering
    \includegraphics[width=.45\textwidth]{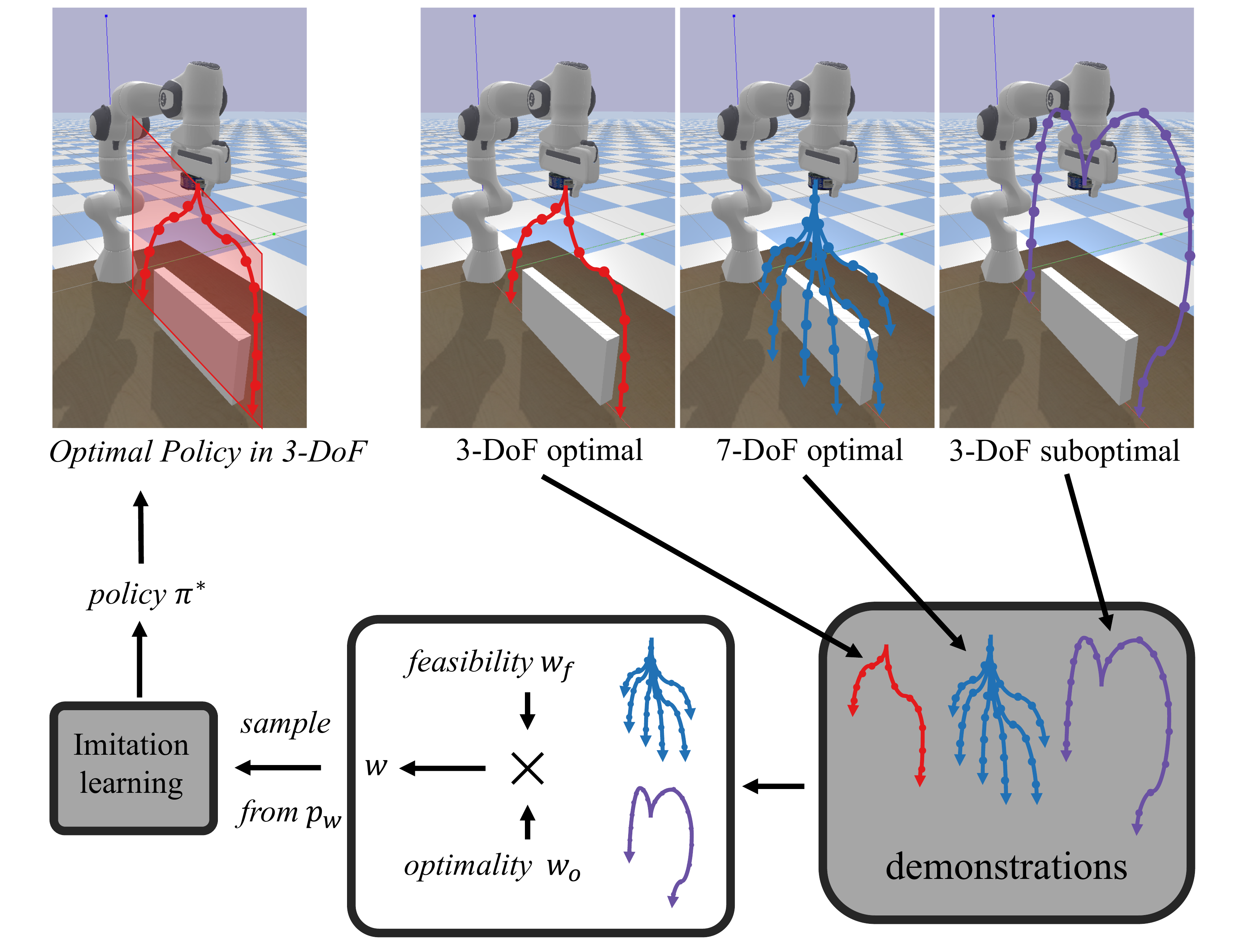}
    \vspace{-10pt}
    \caption{Learning from Imperfect Demonstrations from Agents with Varying Dynamics. The target agent is a 3-DoF robot arm, and the demonstrations can be: optimal demonstrations in the same dynamics (\emph{3-DoF optimal}), optimal demonstrations from another dynamics (\emph{7-DoF optimal}), or suboptimal demonstrations in the same dynamics (\emph{3-DoF suboptimal}). We develop a score $w$ derived from a feasibility $w_f$ and an optimality $w_o$ to define a distribution $p_w$ over state transitions in demonstrations. $p_w$ assigns higher probability to desirable demonstrations to enable learning an optimal policy.}
    \label{fig:pull_fig}
    \vspace{-8pt}
\end{figure}

We introduce an algorithm for learning from agents with varying dynamics. 
Our algorithm leverages a new metric that combines a feasibility score and an optimality score to assess the informativeness of the demonstrations. 
The feasibility score measures the likelihood of a trajectory to be achieved by a target agent.
The optimality score measures optimality under different dynamics as opposed to prior work that only assess optimality based on the demonstrator's dynamics. Leveraging our new metric, we can learn from demonstrations that are both more likely to be feasible and have higher expected return.
The proposed score only changes how we sample from the demonstrations, and our algorithm is agnostic to the choice of imitation learning algorithm used after scoring the demonstrations.

The main contribution of the paper can be summarized as:
\begin{itemize}
    \item We propose a more realistic problem setting of learning from imperfect demonstrations, where the demonstrations may not be optimal and can be drawn from agents with different dynamics.
    \item We propose a scoring function composed of a feasibility score and an optimality score to measure how informative each imperfect demonstration is. The feasibility score measures how likely it is for a state transition to be achieved by the target agent, while the optimality score measures how optimal a state transition is with respect to the expert behavior according to its expected return. 
    \item We conduct experiments in four simulation environments and an obstacle avoidance task on a robot. We show that our algorithm outperforms other methods when learning from imperfect demonstrations with varying dynamics.
\end{itemize}

\section{Related work}

\noindent \textbf{Imitation Learning.} The goal of imitation learning is to find a policy that best imitates expert demonstrations.
Behavior cloning (BC) directly learns the policy from a sequence of state-action pairs~\cite{bain1995framework}.
However, BC can result in compounding errors that is usually addressed by dataset aggregation~\cite{ross2011reduction} or policy aggregation~\cite{daume2009search,ross2010efficient}.
On the other hand, Inverse reinforcement learning (IRL) aims to first recover the reward function from the demonstrations and then find a policy through reinforcement learning~\cite{abbeel2004apprenticeship,ng2000algorithms,ziebart2008maximum,fu2017learning}. 
Generative Adversarial Imitation Learning (GAIL) was proposed for learning the expert behavior by matching the occupancy measure between the policy and the demonstrations~\cite{ho2016generative}.
GAIL has been extended to improve the divergence measurement~\cite{xiao2019wasserstein,ke2019imitation}, but these extensions rely on access to a sequence of states and actions. When only state observations are available, imitation learning first recovers the actions~\cite{torabi2018behavioral}, or directly matches the state distribution~\cite{schroecker2017state,torabi2018generative,sun2019provably}. 

\noindent \textbf{Learning from Demonstrations with Different dynamics.} The above imitation learning works assume the demonstrator shares the same MDP as the target agent.
However, in reality, the demonstrator may have a different dynamics than the target agent, which introduces the correspondence problem between the demonstrator and the target agent~\cite{nehaniv2002correspondence,argall2009survey}. Englert \emph{et al.} address the correspondence problem by aligning the state trajectory distributions of the expert demonstrations and of the policy~\cite{englertaddressing}. Calinon \emph{et al.} model the demonstrations as a Gaussian mixture model within a projected lower-dimensional subspace, and reproduce the optimal behavior by optimizing a time-dependent similarity measurement for different robot contexts~\cite{calinon2007learning}. Eppner \emph{et al.} learn a task description as the relation of objects and robot body parts using a dynamic Bayesian network and reproduce the joint action by maximizing the likelihood~\cite{eppner2009imitation}. Liu \emph{et al.} aim to address the different dynamics for imitation learning by matching the state sequence distribution to learn a state policy and recover the actions using inverse dynamics~\cite{liu2019state}. These works need a correspondence between the demonstrations and the target agent, but under different dynamics, there could exist situations that the demonstrations are not even achievable by the target agent, and the correspondence may not exist. Our feasibility metric is designed to address exactly this problem. Also, our work unlike prior works does not assume access to optimal demonstrations.

\noindent \textbf{Learning from Imperfect Demonstrations.} 
Previous learning from demonstration works usually assume that the demonstrations are provided by an optimal agent who is successful at achieving the task~\cite{codevilla2018end,ho2016generative,ross2011reduction,abbeel2004apprenticeship,fu2017learning}.
In practice, most demonstrations may be sub-optimal or even failures due to reasons, such as the difficulty of providing expert data on robots with high degrees of freedom, or noisy demonstrations due to bounded rationality of humans~\cite{basu2017you,akgun2012keyframe,losey2020controlling,kwon2020humans}.
Recently, several works have addressed the problem of learning from imperfect demonstrations including learning a confidence weight to model the optimality of state-action pairs~\cite{pmlr-v97-wu19a}, or modeling the probability of optimal demonstrations~\cite{tangkaratt2019vild}. Zhu \emph{et al.} utilize a sparse reward function to measure the optimality of each demonstration~\cite{zhu2020learning}. Grollman \emph{et al.} assume access to only low-quality demonstrations, and treat the data as negative examples~\cite{grollman2011donut}. 
All of these works assume the same MDP between the provided demonstrations and the target agent. This results in simplified optimality metrics based on the demonstrator's dynamics. However, when the dynamics of a demonstrator and the target agent are different, the previous optimality measurements are not applicable anymore. Here, we introduce a new optimality measure combined with a feasibility metric to tackle the problem of learning from imperfect demonstrations from agents with varying dynamics.

\section{Problem Statement}

We consider the standard Markov decision process (MDP): $\mathcal{M}= \langle \mathcal{S}, \mathcal{A}, p, \mathcal{R}, \rho_0, \gamma \rangle$, where $\mathcal{S}$ is the state space, $\mathcal{A}$ is the action space, $p: \mathcal{S} \times \mathcal{A} \times \mathcal{S}\rightarrow [0,1] $ is the transition probability, $\rho_0$ is the initial state distribution, $\mathcal{R}: \mathcal{S} \times \mathcal{A} \rightarrow \mathbb{R}$ is the reward function, and $\gamma$ is the discount factor.

A policy $\pi: \mathcal{S} \times \mathcal{A} \rightarrow [0,1]$ defines a probability distribution over the space of actions in a given state. An optimal policy $\pi^*$ maximizes the expected return $\eta_{\pi}=\mathbb{E}_{\xi \sim \pi}\large[G_t(\xi)\large] = \mathbb{E}_{s_0 \sim \rho_0,\pi}\left[\sum_{t=0}^{\infty}\gamma^{t} \mathcal{R}(s_{t}, a_{t}, s_{t+1})\right]$, where $t$ indicates the time step.

We formalize the problem of learning the optimal policy $\pi^*$ for the MDP $\mathcal{M}$ as an imitation learning problem: Given a set of demonstration trajectories $\Xi=\{\xi_1, \dots, \xi_D\}$, where each trajectory is a sequence of state-action pairs $\xi=\{s_0, a_1, s_1, a_2, \dots,a_N, s_N\}$, we aim to learn a policy $\pi^*$ that best matches the demonstration trajectories. Note that we focus on the offline imitation setting as employed in~\cite{ho2016generative,pmlr-v97-wu19a}, where a set of demonstrations are provided ahead of time instead of gradually incrementing our dataset as in online imitation learning~\cite{lee2019continuous}.

One of the core assumptions in prior works in imitation learning is that the demonstrations are drawn from the expert policy of the MDP $\mathcal{M}$, which is referred as the \emph{target} MDP~\cite{ho2016generative,fu2017learning,qureshi2018adversarial}. Here, we are interested in a variant of the imitation learning problem, where we do not make the strong assumption of having access to  \emph{expert} demonstrations. 
Instead, we consider a setting, where the demonstrations can be drawn from either a \emph{sub-optimal policy} or from a \emph{different MDP, $\mathcal{M}^d$,} that has a different dynamics from the target MDP $\mathcal{M}$.
We note that the transition probability of an MDP is affected by changes both in the environment and in the agent's dynamics. The \emph{demonstrator} MDP, $\mathcal{M}^d$, differs from the \emph{target} MDP, $\mathcal{M}$, only in the agent's dynamics.
For example, in a driving environment, we might have demonstrations from a vehicle with different vehicle dynamics that drive in the same environment.
Unlike prior imitation learning work, we can only rely on state-transitions as opposed to state-action sequences since the two MDPs might not share the same transition probabilities, and some actions in the demonstrator's MDP might not even be feasible in the target MDP.
We thus assume the reward is only a function of state transitions, i.e., $\mathcal{R}: \mathcal{S} \times \mathcal{S} \rightarrow \mathbb{R}$, and instead of imitating the state-action trajectories, \emph{we discard actions in the demonstration trajectories and let the agent imitate the state transitions}, i.e., $(s_{t}, s_{t+1})$ pairs.

\noindent \textbf{Challenges.} The core challenge in learning from imperfect demonstrations from agents with varying dynamics is assessing how beneficial a given trajectory is for learning a policy for the target MDP.
This requires measuring two main characteristics of a given trajectory: 1) whether the trajectory is achievable in the target MDP (\emph{feasibility}), and 2) the expected return of the trajectory in the target MDP (\emph{optimality}).

We argue that learning a policy can only benefit from a trajectory if and only if neither the feasibility nor the optimality are low for that given trajectory. 
If a trajectory has low feasiblity, it is not achievable by the target agent and thus the target agent cannot imitate the behavior. Similarly, if a trajectory has low optimality, imitating it can induce a sub-optimal policy leading to low expected return.

\section{Learning from Imperfect Demonstrations from Agents with Varying Dynamics}
In this section, we first introduce our method that incorporates a scoring mechanism for assessing feasibility and optimality, and then discuss our algorithm.

\subsection{Feasibility}
Feasibility measures how likely it is for each state transition in the demonstration to be generated by the target MDP.
Previous learning from imperfect demonstration works do not consider a measure of feasibility since they often assume the demonstrator follows the same dynamics as the target MDP~\cite{pmlr-v97-wu19a,tangkaratt2019vild}.
Formally, a state transition $(s_t,s_{t+1})$ is \emph{feasible} if and only if there exists an action $a_t \in \mathcal{A}$ in the target MDP $\mathcal{M}$ satisfying $p(s_t,a_t,s_{t+1}) > 0$.

However, we cannot simply apply this definition to our demonstrations to verify feasibility. 
First, the transition probability function is usually unknown or difficult to learn in most realistic robotics tasks, and thus we cannot directly use the function to check for the feasible actions. If we do not use the transition probability function, but assume access to a simulator, where we can sample the next state given a state and action -- as is assumed by prior works~\cite{ho2016generative,pmlr-v97-wu19a,fu2017learning,qureshi2018adversarial} -- we would still need to search over a large space of possible actions to verify existence of an action $a_t$ for the state transition $(s_t,s_{t+1})$ to be successful in the target MDP.

We instead propose computing the most probable action using an inverse dynamics model of the target MDP $f_{\text{id}}:\mathcal{S}\times \mathcal{S} \rightarrow \mathcal{A}$, which takes a state transition pair $(s_t, s_{t+1})$ that could be generated by either the target MDP or the demonstrator MDP as an input and outputs all the actions $a_t$ that satisfy $p(s_t,a_t,s_{t+1})>0$.
For every $\xi$ from the set of demonstrations provided by the demonstrator MDP, we initialize the agent at the state $s_0$, which is the initial state of $\xi$. We then make the agent take an action $a'_t$ generated by $f_{\text{id}}$ at each time step to generate a new state trajectory $\xi'=\{s'_0, s'_1, s'_2 \dots, s'_N \}$ that is achievable by the target MDP:
\begin{equation}\label{eqn:id}
\begin{aligned}
    & s'_0 = s_0, \quad a'_t = f_{\text{id}}(s'_{t-1}, s_{t}) , t\ge1 \\
    & s'_{t} \sim p(s'_{t-1}, a'_t, s'_{t}), t\ge1\\
\end{aligned}
\end{equation}
\noindent \textbf{Learning Inverse Dynamics.} If we assume access to an accurate inverse dynamics model $f_{\text{id}}$ -- meaning that $f_{\text{id}}$ gives the correct action for a feasible state transition and outputs ``None'' for an infeasible state transition, we can detect the infeasible trajectories based on the ``None'' output.
However, such a binary feasibility metric cannot differentiate between nearly feasible and totally infeasible trajectories, where the former is useful for learning and the latter does not have much learning value.
So instead we aim to learn a $f_{\text{id}}$ that outputs the correct action if the state transition is feasible while still outputting the closest action, when the state transition is infeasible.
We model $f_{\text{id}}$ as a neural network, and train it on a randomly sampled set of trajectories $\Xi_f=\{\xi_f\}$ from the target MDP. We emphasize that similar to prior imitation learning works~\cite{ho2016generative,pmlr-v97-wu19a,fu2017learning,qureshi2018adversarial}, we assume access to a simulator for the target MDP, where we can sample feasible but random trajectories. We learn $f_{\text{id}}$ using a regression loss:
\begin{equation}
    \min_{f_{\text{id}}} \mathbb{E}_{(s_{t}, a, s_{t+1})\sim \xi_f, \xi_f \sim \Xi_f} L_{1;\text{smooth}}(f_{\text{id}}(s_{t},s_{t+1}), a).
\end{equation}
\noindent \textbf{Computing the Feasibility Score.} Here, the trained $f_{\text{id}}$ will output similar actions if the input state transition is similar to a state transition seen in the training data.
With the learned $f_{\text{id}}$, for every trajectory $\xi$ -- even if it is infeasible -- we can compute a corresponding trajectory $\xi'$.
In addition, we can compute a distance metric between the two trajectories $\xi$ and $\xi'$, $F(\xi, \xi')$, where lower distances correspond to higher feasibility scores. We compute the feasibility score by normalizing this distance within a range $[d_{\text{min}}, d_{\text{max}}]$:
\begin{equation}\label{eqn:feasibility}
    w_f(\xi) = \begin{cases}
    1 & F(\xi, \xi') < d_{\text{min}} \\
    1- \frac{F(\xi, \xi')-d_{\text{min}}}{d_{\text{max}}-d_{\text{min}}} & d_{\text{min}}\le F(\xi, \xi') \le d_{\text{max}}\\ 
    0 & F(\xi, \xi') > d_{\text{max}} \\
    \end{cases} 
\end{equation}
Here, $\xi'$ is a corresponding trajectory to $\xi$ generated by the inverse dynamics $f_{\text{id}}$.
Further, we can choose $F(\xi, \xi')$ to be any sequence metric between the two trajectories. In our experiments, we computed the mean of the $l_2$ distance between each pair of states in $\xi$ and $\xi'$. The proposed continuous feasibility score can measure different degrees of feasibility and preserve nearly feasible trajectories that can be useful for learning a policy for the target agent.

\noindent \textbf{Discussion of Feasibility Score.} Using our feasibility measurement, the more infeasible transitions contained in $\xi$ are, the deviation between $\xi'$ and $\xi$ becomes larger, and hence it would be less likely that the trajectory is drawn from the target MDP.
A possible drawback of our feasibility score is compounding errors from our inverse dynamics model $f_{\text{id}}$.
To address this problem, we normalize the distance between the trajectories $F(\xi, \xi')$ using $d_{\text{min}}$ and $d_{\text{max}}$. Here, $d_{\text{min}}$ models the lowest compounding error that could be achieved by a feasible trajectory, while $d_{\text{max}}$ is the highest compounding error. Thus, our feasibility measure can assign non-zero feasibility to any feasible trajectories even with compounding errors from the inverse dynamics. We include the method to set the thresholds in Section II of the supplementary materials.

\subsection{Optimality}\label{sec:optimality}
We would like to define an optimality score that not only measures the instance reward received by the state transition but also the influence of a given state transition on the future state transitions, e.g., whether the given state transition leads to higher expected return trajectories.
Previous works evaluate optimality based on pre-defined measurements provided by the demonstrator, such as confidence score~\cite{pmlr-v97-wu19a} and probability of taking an optimal action~\cite{tangkaratt2019vild}.
However, these scores are evaluated based on the demonstrator's dynamics. 

\begin{figure}[ht]
    \centering
    \subfigure[Different Dynamics]{\includegraphics[width=0.2\textwidth]{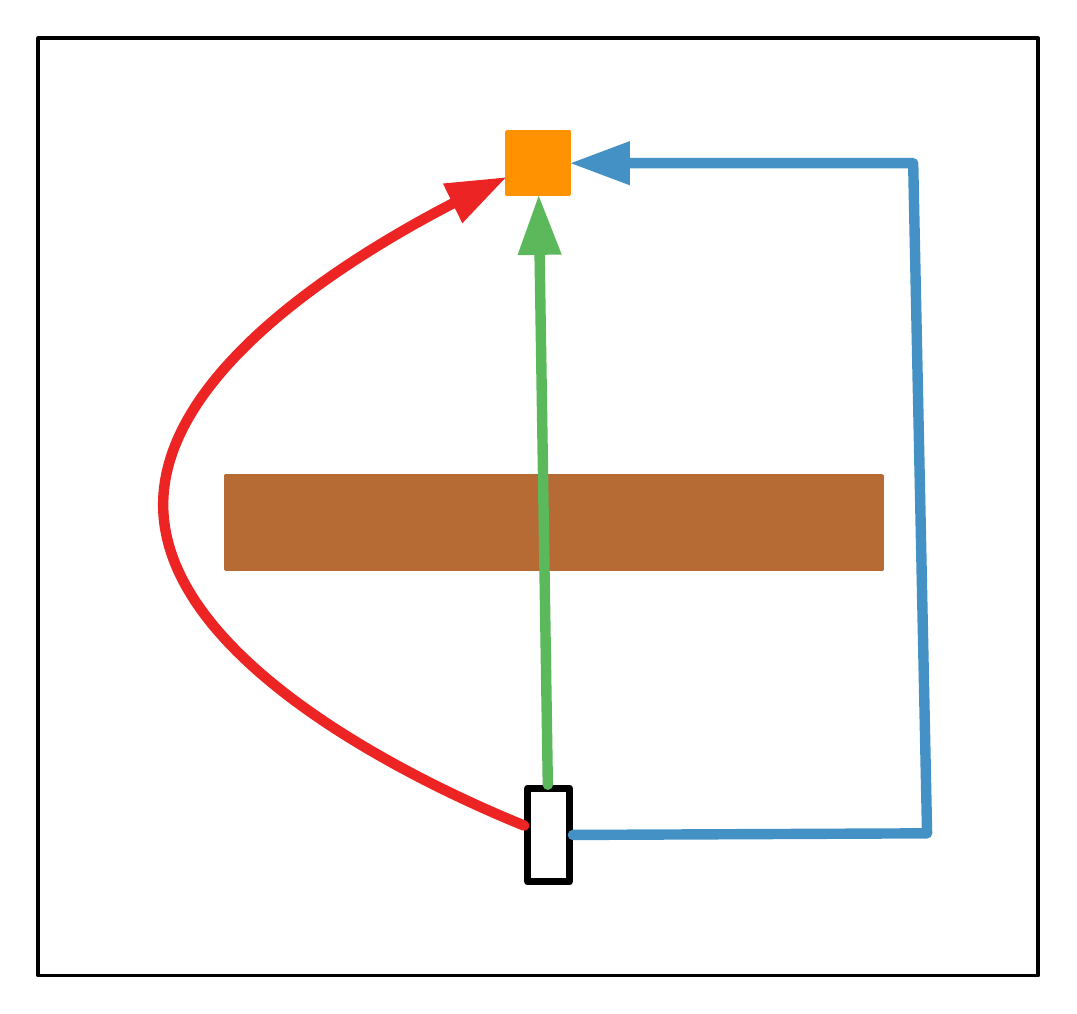}\label{fig:dd}}
    \hfil
    \subfigure[Different Initial States]{\includegraphics[width=0.2\textwidth]{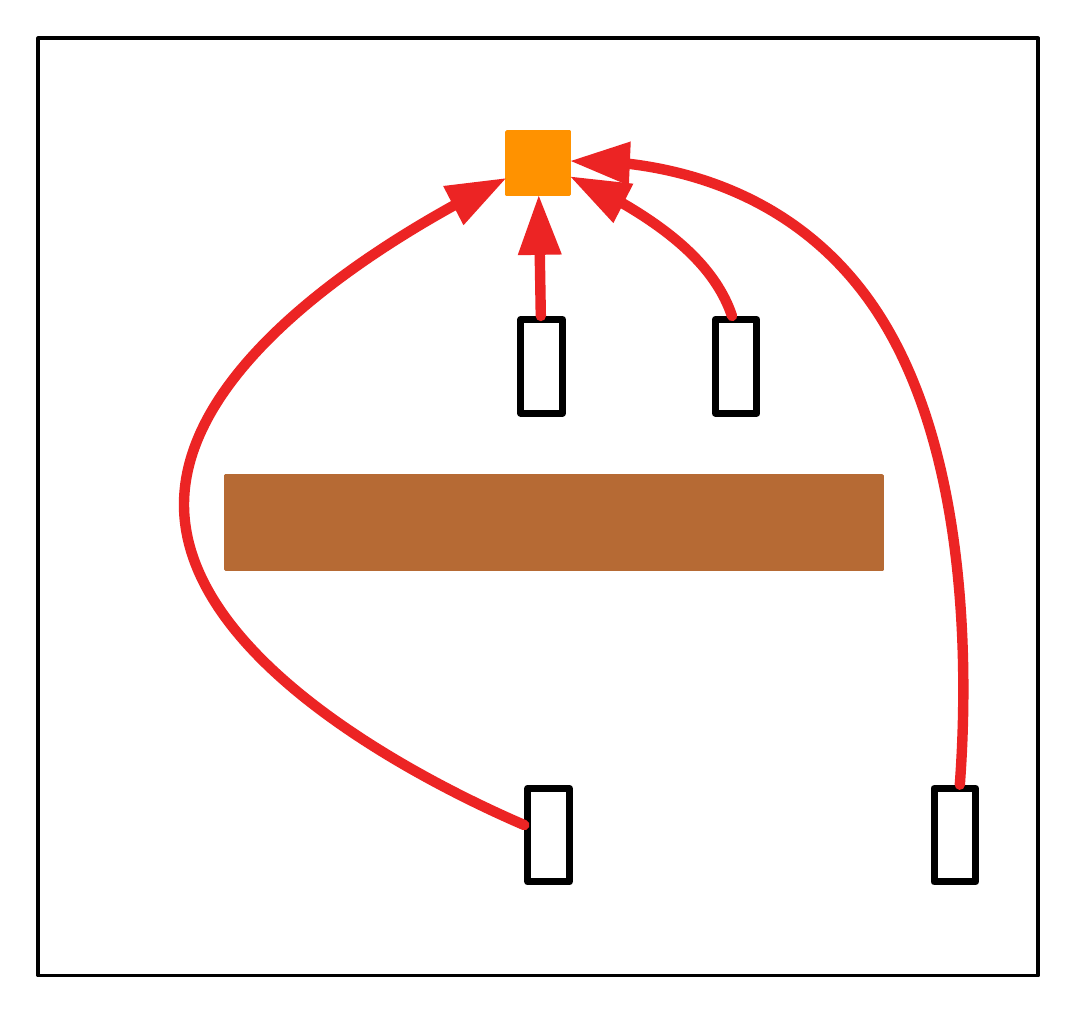}\label{fig:dis}}
    \vspace{-10pt}
    \caption{Fig. (a) shows optimal trajectories of navigation to the orange goal without colliding with the brown obstacle under different dynamics. The green trajectory is for a car that can never turn and only go straight. The blue trajectory is for a car that can only turn in $\frac{\pi}{2}$ increments. The red trajectory is for a car that can turn in any direction. We can see the optimal trajectories for different dynamics under the same reward can be different. Fig. (b) shows the optimal trajectories from different initial states with the same dynamics under the navigation task to reach the orange goal as fast as possible. So the car receives positive reward for reaching the goal but is punished with negative reward for every time step. The expected returns received by the optimal trajectories are different for different initial states.}
    \label{fig:traj}
    \vspace{-5pt}
\end{figure}

Here, the demonstrator follows a different dynamics as the target agent, and thus the set of state transitions achievable by the target agent may be different from those achievable by the demonstrator. 
For example, consider a navigation setting in Fig.~\ref{fig:dd}, where we allow three possible dynamics: (1)~unlimited turns, (2)~no turns are allowed, and (3)~can only turn in $\frac{\pi}{2}$ increments. 
We set the reward as driving to the orange goal as fast as possible without colliding with the brown obstacle.
Here, the red, green and blue trajectories are the optimal trajectories achieved by the dynamics of \emph{unlimited turns}, \emph{no turns allowed}, and \emph{only turning in $\frac{\pi}{2}$ increments} respectively.
Note that the trajectory with \emph{unlimited turns} receives higher expected return than the other trajectories.
So a state transition that is optimal under one dynamics may be suboptimal or even useless for another agent.
Thus, we need an optimality measurement that can handle the more general case of learning from demonstrations with varying dynamics.

We compute the expected return received by each state trajectory $\xi=\{s_{1},s_{2},\dots,s_N\}$: $\eta_{\xi} = \sum_{t}\gamma^t\mathcal{R}(s_{t},s_{t+1})$ as an estimate for how optimal a trajectory is. 
Since the reward function only depends on state transitions, the expected return of a trajectory can be naturally generalized across different dynamics.
We emphasize that we do not assume access to the reward function, and only assume access to the expected return of a trajectory similar to prior work that leverage a given confidence measure for each demonstration~\cite{pmlr-v97-wu19a}.

\noindent \textbf{Effects of Initial States on Optimality.}
The expected return is general but it still suffers from some problems. 
Is a trajectory with higher expected return really more useful for the policy learning of the target agent?
If the trajectories are starting from the same initial state, this claim is correct, because higher expected return means we have a more optimal path from this initial state.
However, when the trajectories start from different initial states, the claim may not be correct.
Imagine the case shown in Fig.~\ref{fig:dis}, where the car navigates to the goal from different initial states. 
If the car is close to the goal, e.g., one step from the goal, the navigation seems easier while if the car is far away from the goal, the navigation is more difficult.
The optimal trajectory for the close initial location receives much higher reward than the optimal trajectory for the far one and thus is assigned with a much higher optimality if we only use the expected return as our measure of optimality. 
However, both trajectories (starting close or far) are important for the target agent since they give the agent guidance on how to navigate to the goal from different initial locations. 
In fact, we should assign similar optimalilty to both trajectories.
Therefore, na\"ive optimality scores based on the expected return may negatively influence the diversity of the demonstrations. 

\noindent \textbf{The Optimality Score.} To address this problem, we rectify the expected return to make it conditioned on the initial states.
Our key idea is that if demonstration trajectories are from the same initial states, we should learn from the one with higher expected return.
Based on this idea, we define a function $f_{\text{rec}}: \mathcal{S}\rightarrow \mathbb{R}$. The input is an initial state $s_0$ and the output is the highest expected return of a demonstration starting from $s_0$. We then define the optimality of a trajectory $\xi$ as:

\begin{equation}\label{eqn:sigma}
    w_o(\xi) = \exp\left(-\frac{\left(\eta_{\xi} - f_{\text{rec}}(s_0)\right)^2}{2\sigma^2}\right).
\end{equation}
We compute the distance between the expected return of each trajectory $\xi=\{s_0,s_1,\dots, s_N\}$ and the best expected return the demonstration can achieve by starting from $s_0$, and this can determine the optimality score assigned to a trajectory. However, the distance $\eta_{\xi} - f_{\text{rec}}(s_0)$ is smaller than 0 but the lower bound depends on the reward function, which may have highly different scales from the feasibility. So we normalize this score in the range of $[0,1]$. We use a Gaussian function for normalization instead of the widely-used min-max normalization~\cite{zhang2018importance}, which can more effectively filter out extremely low return trajectories.

\noindent \textbf{Learning the Rectifying Function.} The key to the designed optimality is estimating $f_{\text{rec}}$ -- the highest expected return for an initial state.
Given that we only have access to a small number of trajectories, where the initial states and the expected returns are known, we need a set of assumptions for generating training data to learn $f_{\text{rec}}$.
For example, if the initial states of two trajectories $\xi_1$ and $\xi_2$ are very close, the policy learned from one trajectory can also perform on the other and achieve similar expected return. 
So we may assume that $f_{\text{rec}}$ of the two close initial states should be the same or at least similar. 
A realistic assumption for $f_{\text{rec}}$ is the neighborhood property: $\exists \delta>0, \forall \xi=\{s_0,s_1,\dots, s_N\}$, we set the highest expected return for $s_0$ as:

\begin{equation}\label{eqn:frec}
    f_{\text{rec}}(s_0) = \max_{\xi' \in \Xi, \lvert s'_0-s_0 \rvert <\delta,w_f(\xi)>0} \eta_{\xi'},
\end{equation}
where $s'_0$ is the initial state of $\xi'$ and $\Xi$ is the set of all demonstrations.
We set the highest expected return for a initial state as the highest expected return of neighboring initial states of feasible trajectories ($w_f(\xi)>0$), because the rectify function will be influenced by infeasible trajectories with high expected returns otherwise. 
Using this rectifying function $f_{\text{rec}}$, we can compute the rectified optimality, which induces more diverse optimal demonstrations.

\begin{figure*}[ht]
    \centering
    \subfigure[Reacher Clockwise]{\includegraphics[width=0.23\textwidth]{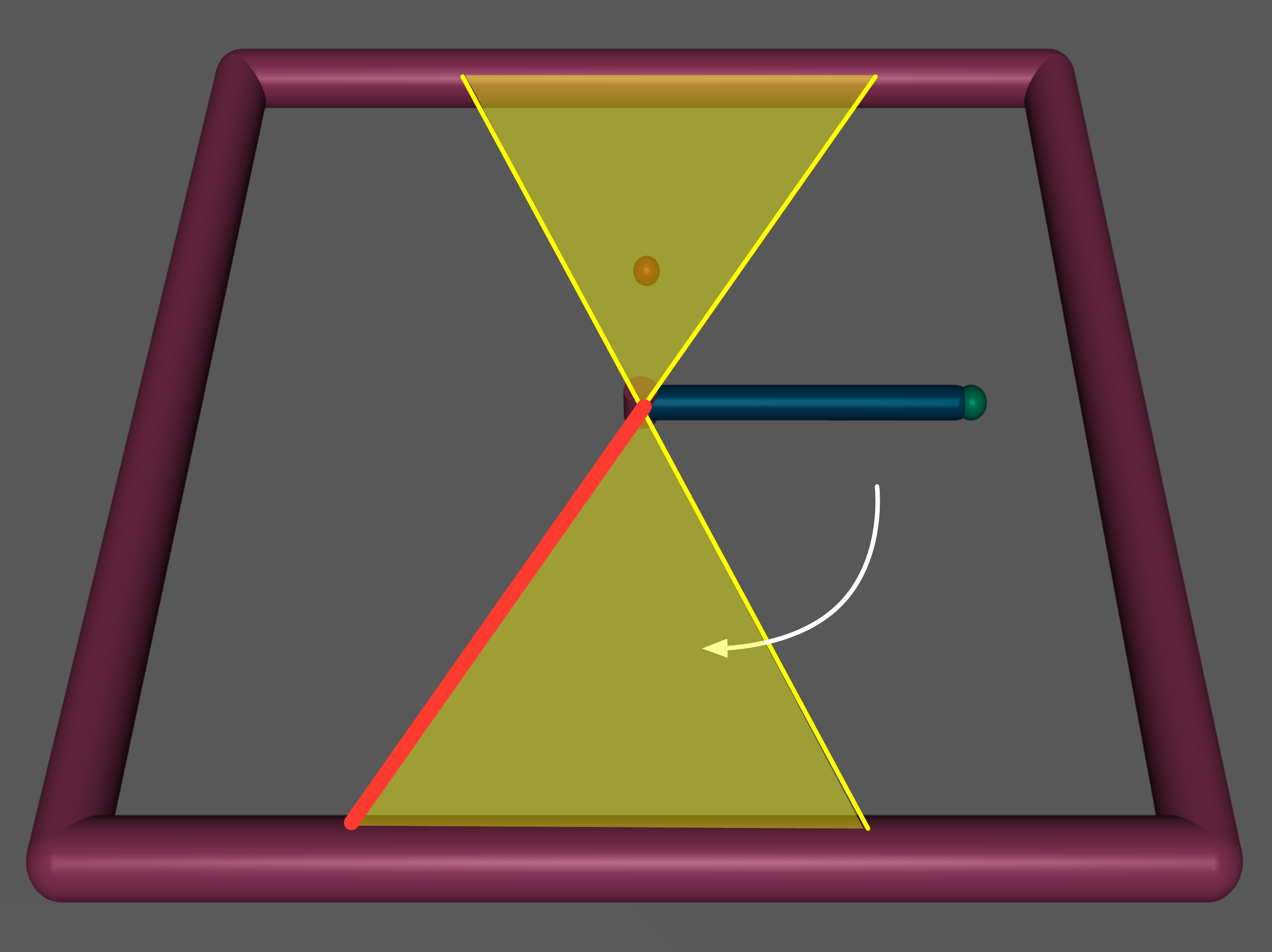}\label{fig:reacher_clock_env}}
    \subfigure[Reacher Counter Clockwise]{\includegraphics[width=0.23\textwidth]{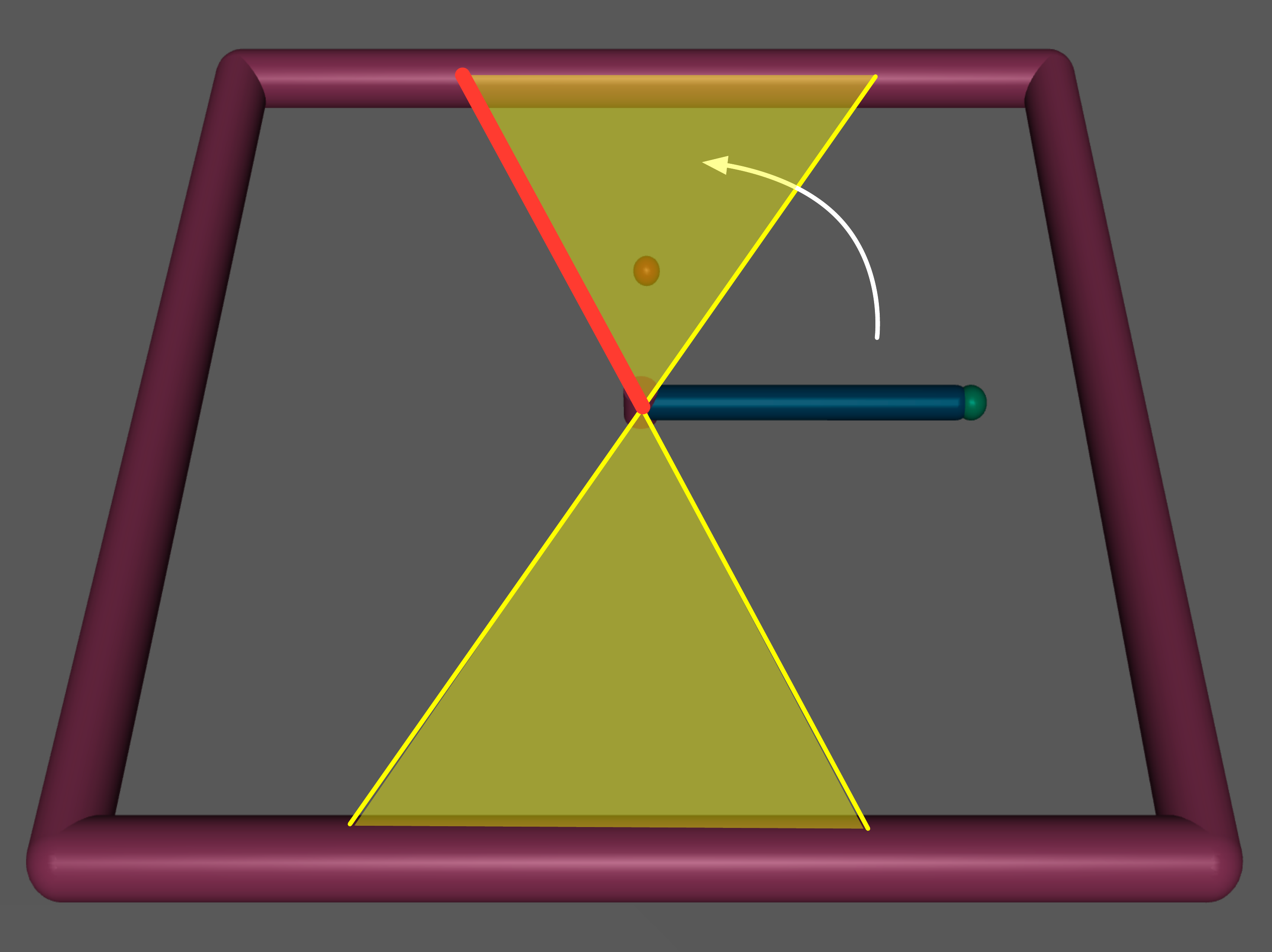}\label{fig:reacher_counter_env}}
    \subfigure[Driving Slow]{\includegraphics[width=0.23\textwidth]{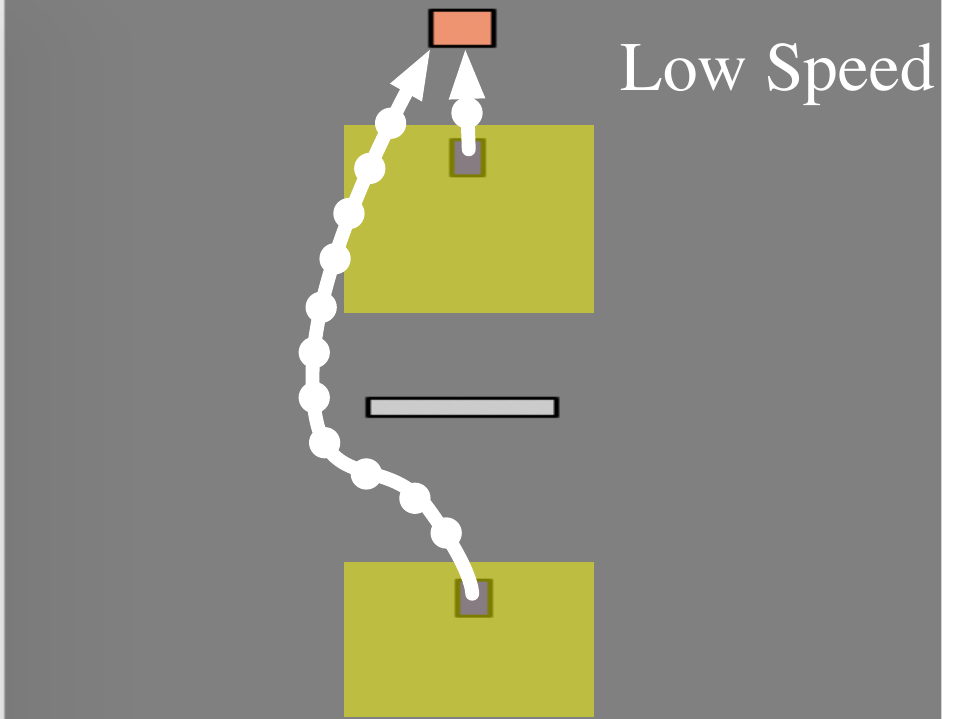}\label{fig:driving_slow_env}}
    \subfigure[Driving Fast]{\includegraphics[width=0.23\textwidth]{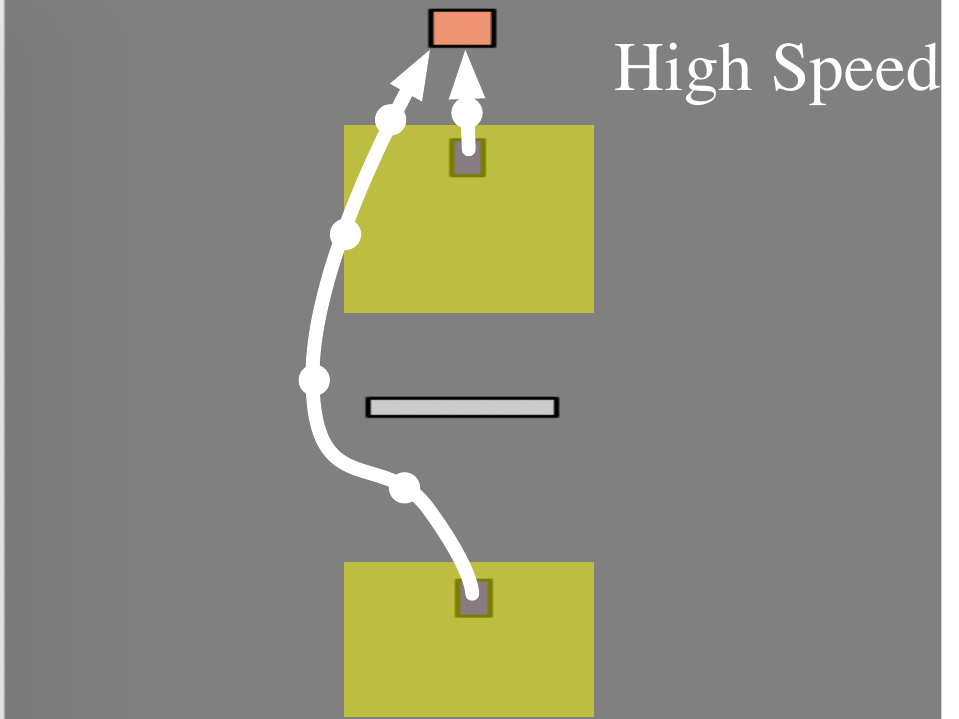}\label{fig:driving_fast_env}}
    \subfigure[Reacher Clockwise]{\includegraphics[width=0.23\textwidth]{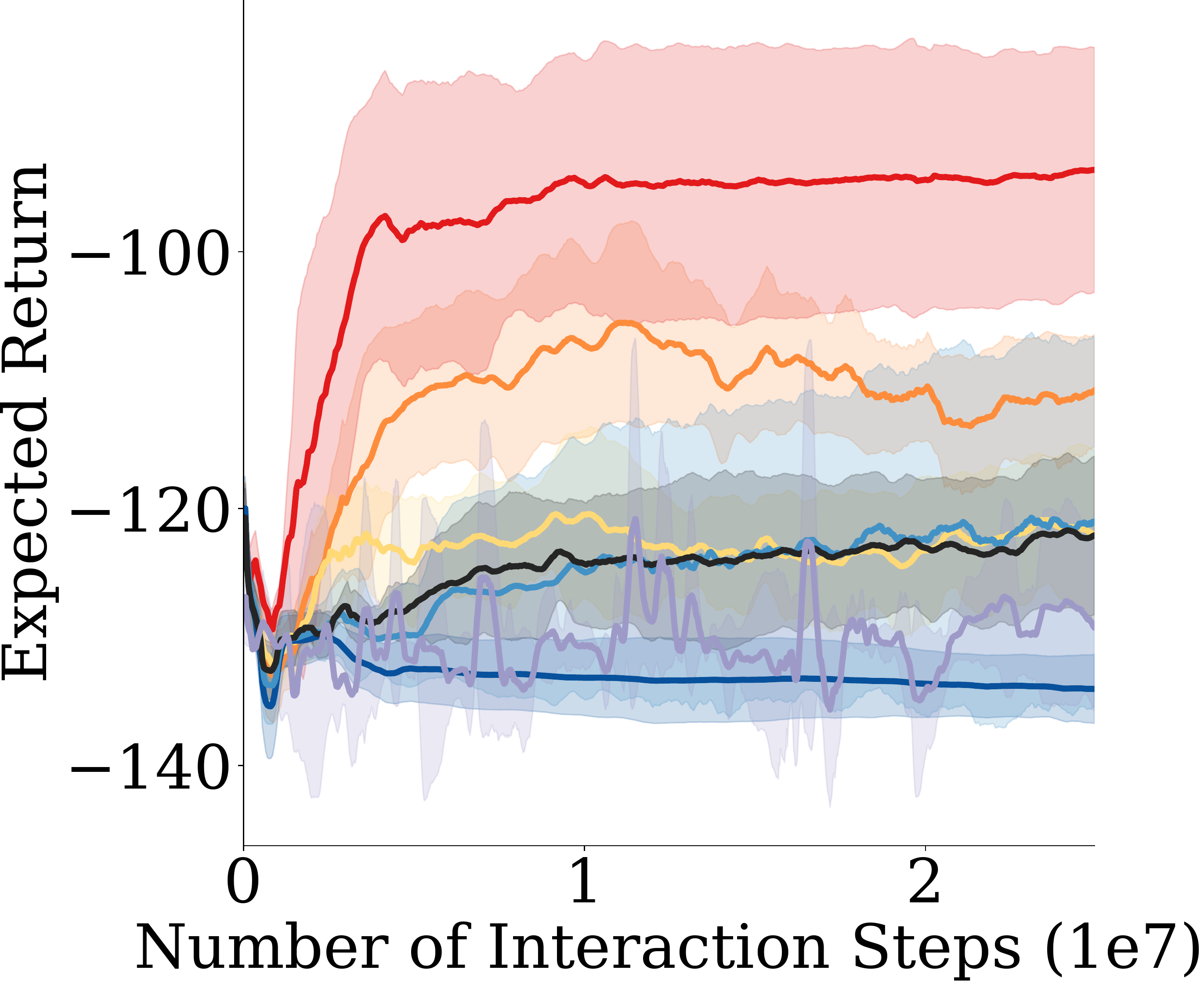}}
    \subfigure[Reacher Counter Clockwise]{\includegraphics[width=0.23\textwidth]{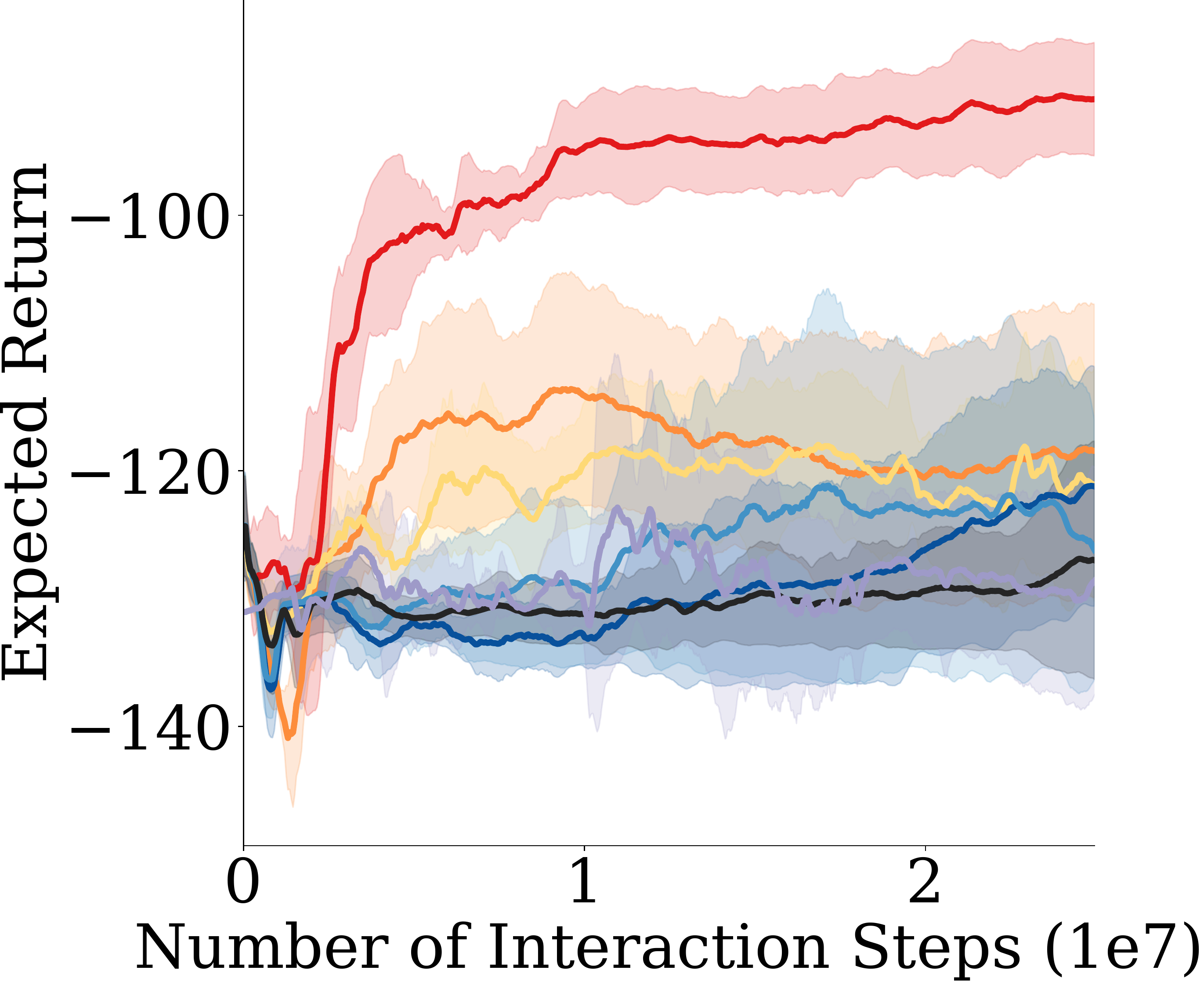}}
    \subfigure[Driving Slow]{\includegraphics[width=0.23\textwidth]{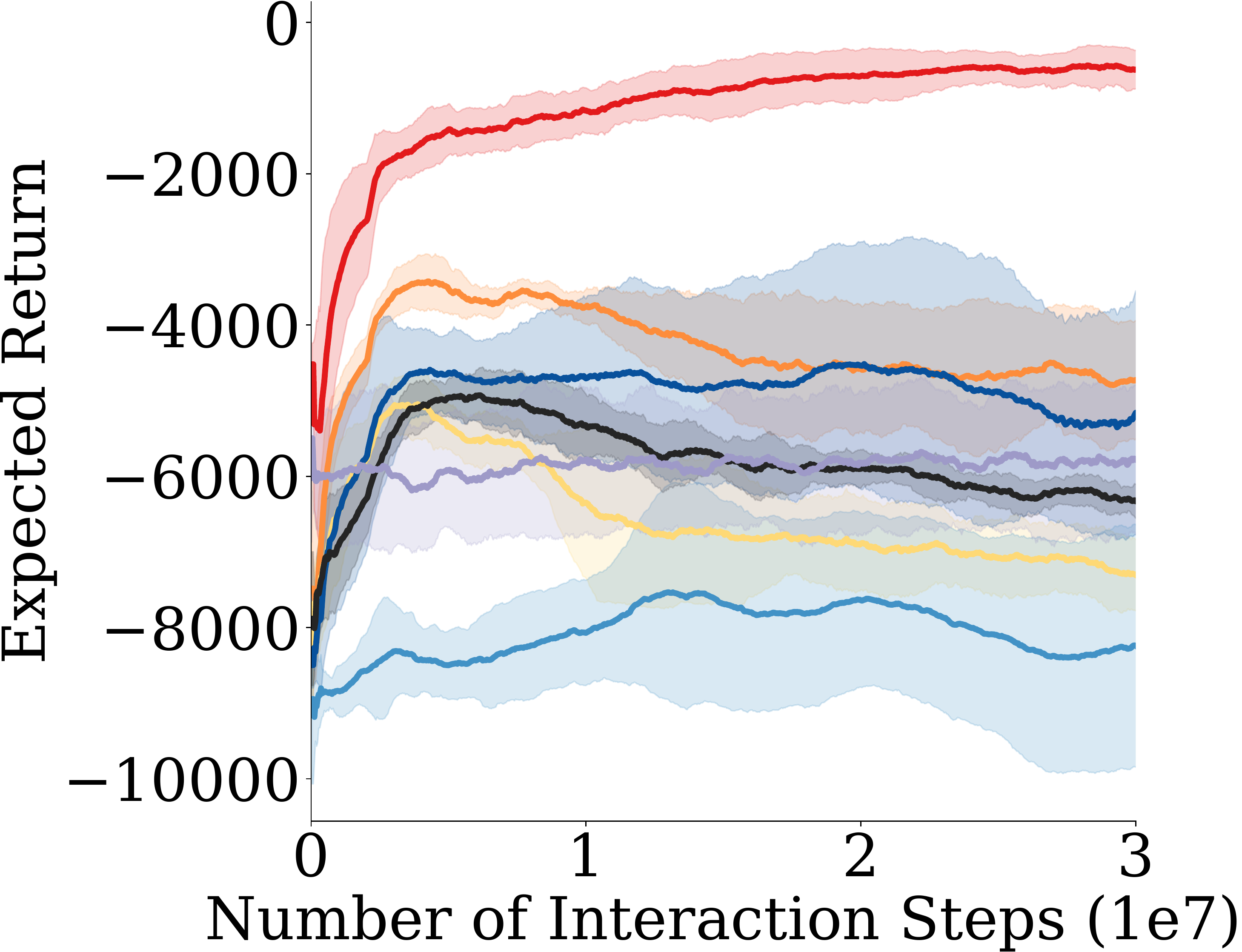}}
    \subfigure[Driving Fast]{\includegraphics[width=0.23\textwidth]{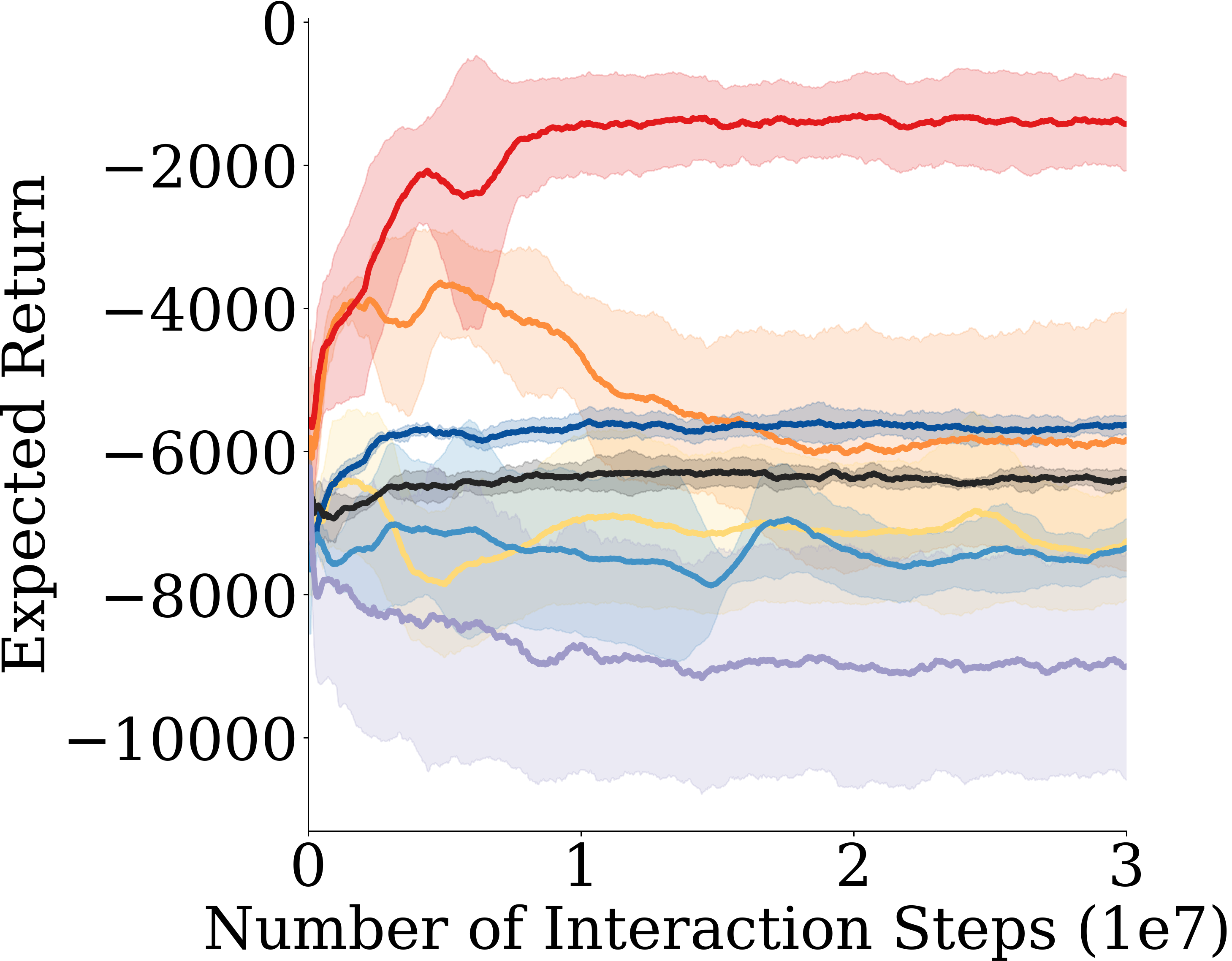}}
    \subfigure{\includegraphics[width=0.9\textwidth]{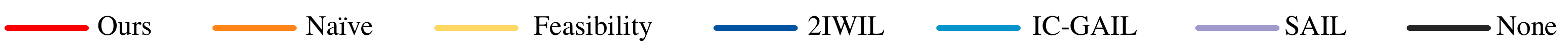}}
    \vspace{-10pt}
    \caption{(a-b) are the illustrations of Reacher environments with clockwise and counter-clockwise dynamics. (c-d) are the illustrations of Driving environments with low speed and high speed dynamics. (e,f,g,h) are the expected return results on environments (a,b,c,d) respectively. One x-axis unit is $10^7$ steps.}
    \label{fig:result_reacher_driving}
    \vspace{-15pt}
\end{figure*}

\subsection{Algorithm}\label{sec:algo}
Using the defined notions of feasibility and optimality, we now can compute the final score of how useful each state transition is for learning a policy for the target agent. 
We compute the final measurement as the product of the feasibility and optimality scores:
\begin{equation}\label{eqn:final_score}
    w(\xi) = w_f(\xi) \times w_o(\xi).
\end{equation}
We then re-weigh each state transition in the demonstration by the score $w(\xi)$. The weighting-based work for learning from imperfect demonstrations often directly incorporate the weight in the imitation loss~\cite{pmlr-v97-wu19a}.
However, such re-weighing can suffer from numerical issues, when the weight of some state transitions are too large. 
Instead, for a more stable training, we define a distribution $p_w$ over state transitions based on the score $w(\xi)$. For every state transition in a demonstration from the dataset, we assign a weight $w_i$ (for the $i$-th state transition) equal to the weight assigned to the trajectory $w(\xi)$, i.e., $\forall i \quad w_i = w(\xi)$. 
We compute a probability distribution over the state transitions as $p_{w_i}=\frac{w_i}{\sum_{i}w_i}$. 
Using the sampling distribution $p_w$, we can embed our method into any imitation learning algorithm to enable it to learn from imperfect demonstrations from various dynamics. 

Note that given a trajectory, we assign equal scores to each state-transition within the trajectory, which fails to emphasize the most important transitions. 
However, this is not a problem as the key transitions repeatedly appear in high score trajectories, and everytime they appear, their sampling probability $p_w$ accumulates. This ensures the key transitions are considered. The detailed description of the algorithm is shown in Section I of the supplementary materials.

\section{Experiment}
We conduct experiments in five environments including three MuJoCo environments: Reacher, Swimmer, and Ant, one driving, and a Robot Arm environment, which are all kinematic without force interactions. However, our approach is not specific to kinematics, and can also be applied to settings with force interactions as long as we can interact with the environment to generate trajectories. We include the implementation details, parameter selection, and results on varied demonstration ratios in the supplementary materials.

\noindent \textbf{Source of Demonstrations.} We collect imperfect demonstrations using a policy trained with only $10$ epochs of reinforcement learning, i.e., the policy is not converged. These generated trajectories range from random trajectories to near-optimal trajectories. We also collect optimal demonstrations with an optimal policy. 
For Reacher, Swimmer, and Ant environments, we train the optimal policy by TRPO~\cite{schulman2015trust}. For Driving and Panda Robot Arm, it is difficult for TRPO to learn an optimal policy, so we hand-craft an optimal policy.

\noindent \textbf{Composition of Demonstrations.} We follow the experiment setting of the state-of-the-art weighting-based learning from imperfect demonstration work~\cite{pmlr-v97-wu19a}. We do not make extra assumptions on the quality of demonstrations, and consider the setting, where demonstrations can range from random to fully optimal. Therefore, we employ a mixture of perfect and imperfect demonstrations, 
where some demonstrations are useful and informative, while others are not as useful or informative. Specifically, we construct a mixture of three types of data: demonstrations of the optimal policy of the target environment, demonstrations of the sub-optimal policy of the target environment, and the demonstrations from an agent with different dynamics.

In each experiment, we use a set of $1000$ demonstrations for each environment.
To achieve a non-trivial imitation learning problem, we select ratios of each type of demonstration appropriately to ensure that the problem cannot easily be solved by vanilla imitation learning. The specific composition of demonstrations in each environment is discussed below.

\noindent \textbf{Other Methods.} We compare our proposed approach with the most relevant learning from demonstration works: imitation learning (specifically we compare with Generative Adversarial Imitation Learning (GAIL))~\cite{ho2016generative}, imitation learning from imperfect demonstration: 2IWIL and IC-GAIL~\cite{pmlr-v97-wu19a}, and imitation learning from different dynamics: SAIL~\cite{liu2019state}.

\subsection{Environments with Return Influenced by the Initial State}
\noindent \textbf{Reacher.} We experiment with openAI gym's reacher environment~\cite{brockman2016openai,todorov2012mujoco}, where the end effector of the agent is supposed to reach a final location.
We modify the original reacher environment, where the agent only has one joint and one link as shown in Fig.~\ref{fig:reacher_clock_env} and~\ref{fig:reacher_counter_env}.
In addition, we limit the final goal location (the red ball) within a particular area, which is shown as the shaded yellow area composing of two triangles, where the vertex angle of the triangle at the center is $53.1^\circ  (2\arctan(0.5))$. 
Here, we consider two possible dynamics: (1) the agent can only rotate clockwise with its joint limit reaching at most the red line shown in Fig.~\ref{fig:reacher_clock_env} and (2) the agent can only rotate counter-clockwise with its joint limit reaching at most the red line shown in Fig.~\ref{fig:reacher_counter_env}.
We use the shaded yellow area and the limit on rotation to make the optimal trajectories for one dynamics to be quite different from the other.
For instance, for a goal location in the up triangle, the optimal trajectories are quite different: for the first dynamics the optimal trajectory is to stay still, while the optimal trajectory for the second dynamics is to rotate counter-clockwise toward the goal. So it is difficult for agents with different dynamics to learn from each other.

\noindent \textbf{Driving.} We consider a simple 2D driving environment. 
As shown in Fig.~\ref{fig:driving_slow_env} and~\ref{fig:driving_fast_env}, the blue car aims to drive to the goal while avoiding the white obstacle. 
As introduced in Section~\ref{sec:optimality}, when different initial states induce different highest expected returns, the optimality based on the raw expected return may fail.
To test the difficulty, we set the initial states to be in the yellow areas and define the reward function (just for evaluation) as $\mathcal{R}(s_{t},s_{t+1})=\alpha_1 \mathbf{1}_{\text{goal}}[s_{t+1}]+\alpha_2 \mathbf{1}_{\text{obstacle}}[s_{t+1}]+\alpha_3 \mathbf{1}_{\text{out}}[s_{t+1}]+\alpha_4$, where $\mathbf{1}_{\text{goal}}$ is an indicator of when the next state $s_{t+1}$ reaches the goal,  $\mathbf{1}_{\text{obstacle}}$ is an indicator of when $s_{t+1}$ collides with the obstacle, and $\mathbf{1}_{\text{out}}$ is an indicator of when $s_{t+1}$ moves out, and $\alpha_i$ are appropriate weights for each indicator.
Then the initial location closer to the goal has higher optimal expected return. 
Here, we fix the car speed and only control the steering of the car. 
As shown in Fig.~\ref{fig:driving_fast_env} and Fig.~\ref{fig:driving_slow_env}, we create different dynamics by setting a high and a low speed for the car.

\noindent \textbf{Compostion of Demonstrations.} The ratio of demonstrations from the optimal policy of the target environment, the sub-optimal policy of the target environment, and the optimal policy in a different environment is $5\%,90\%,5\%$ respectively for the Reacher environment and $30\%,60\%,10\%$ respectively for the Driving environment.

\begin{figure*}[ht]
    \centering
    \subfigure[Swimmer BackDisabled]{\includegraphics[width=.23\textwidth]{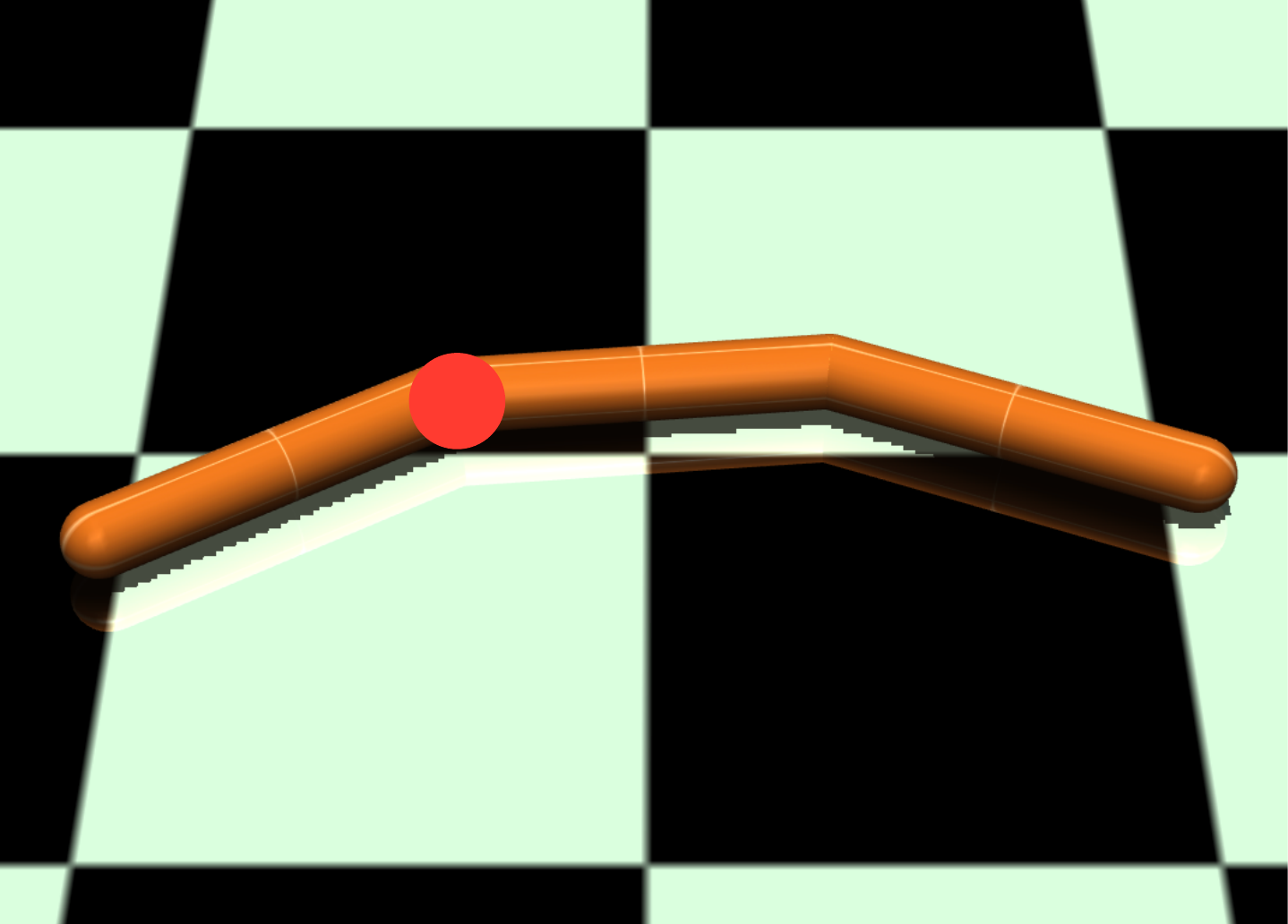}\label{fig:swimmer_back_env}}
    \subfigure[Swimmer FrontDisabled]{\includegraphics[width=.23\textwidth]{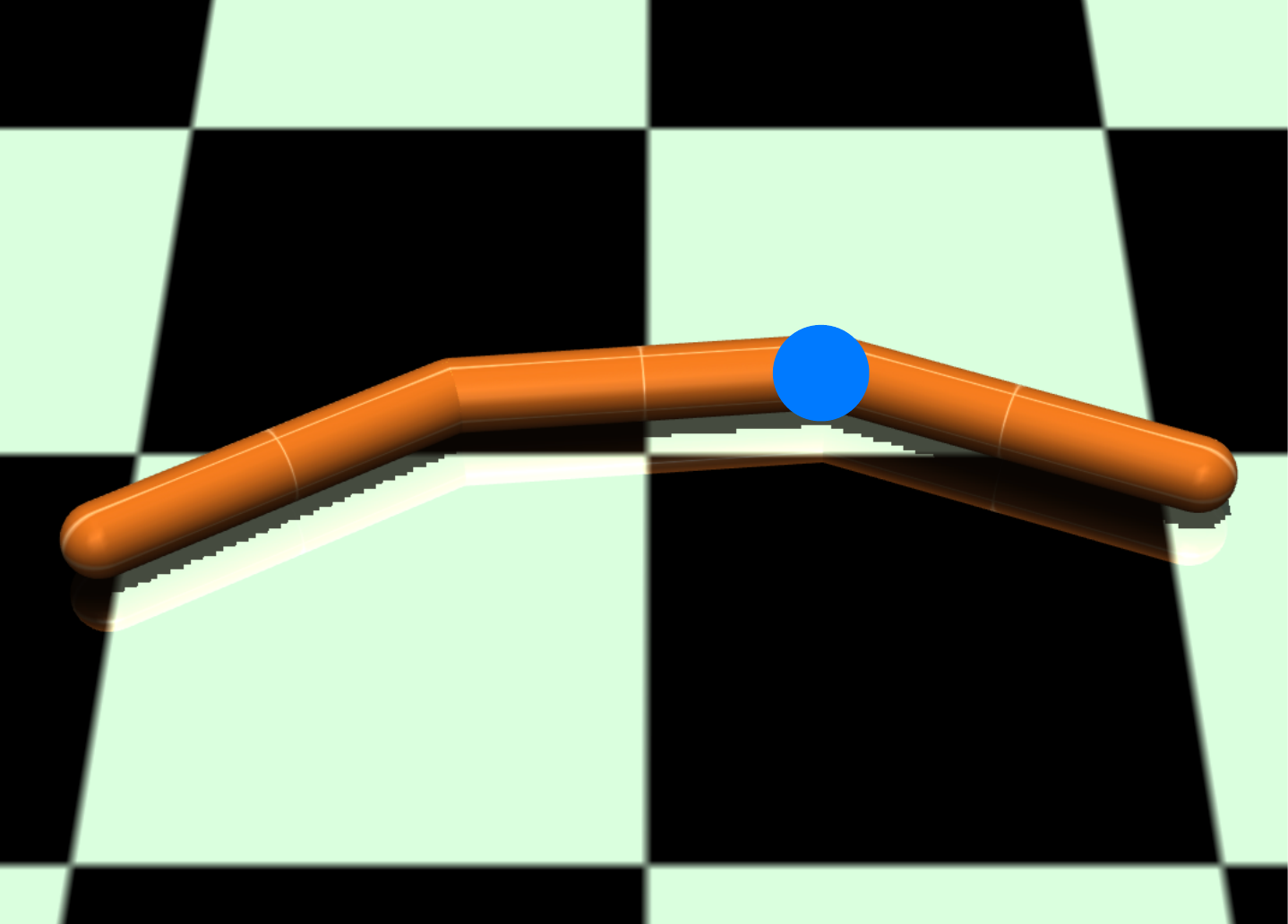}\label{fig:swimmer_front_env}}
    \subfigure[Ant X-axis]{\includegraphics[width=.23\textwidth]{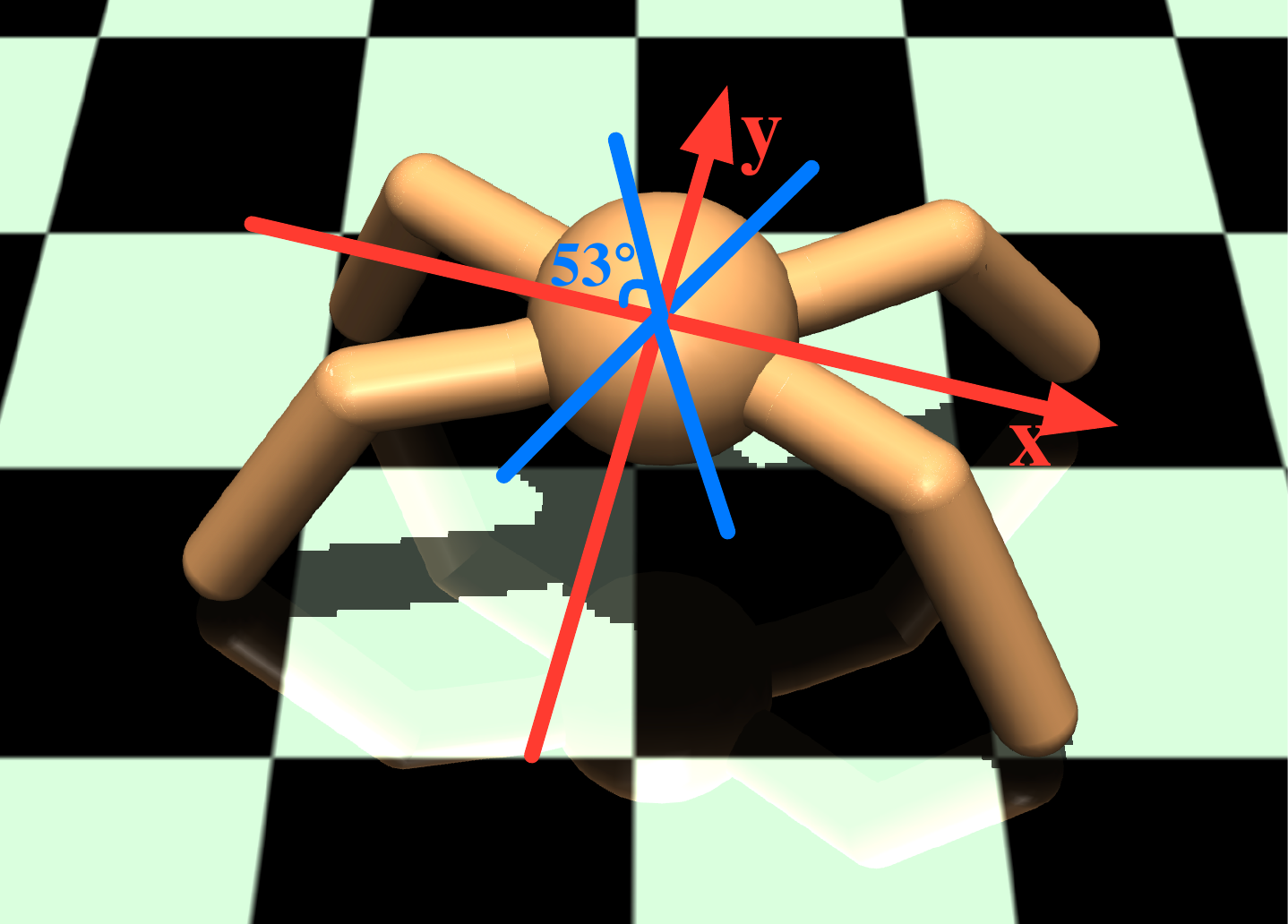}\label{fig:ant_x_env}}
    \subfigure[Ant Y-axis]{\includegraphics[width=.23\textwidth]{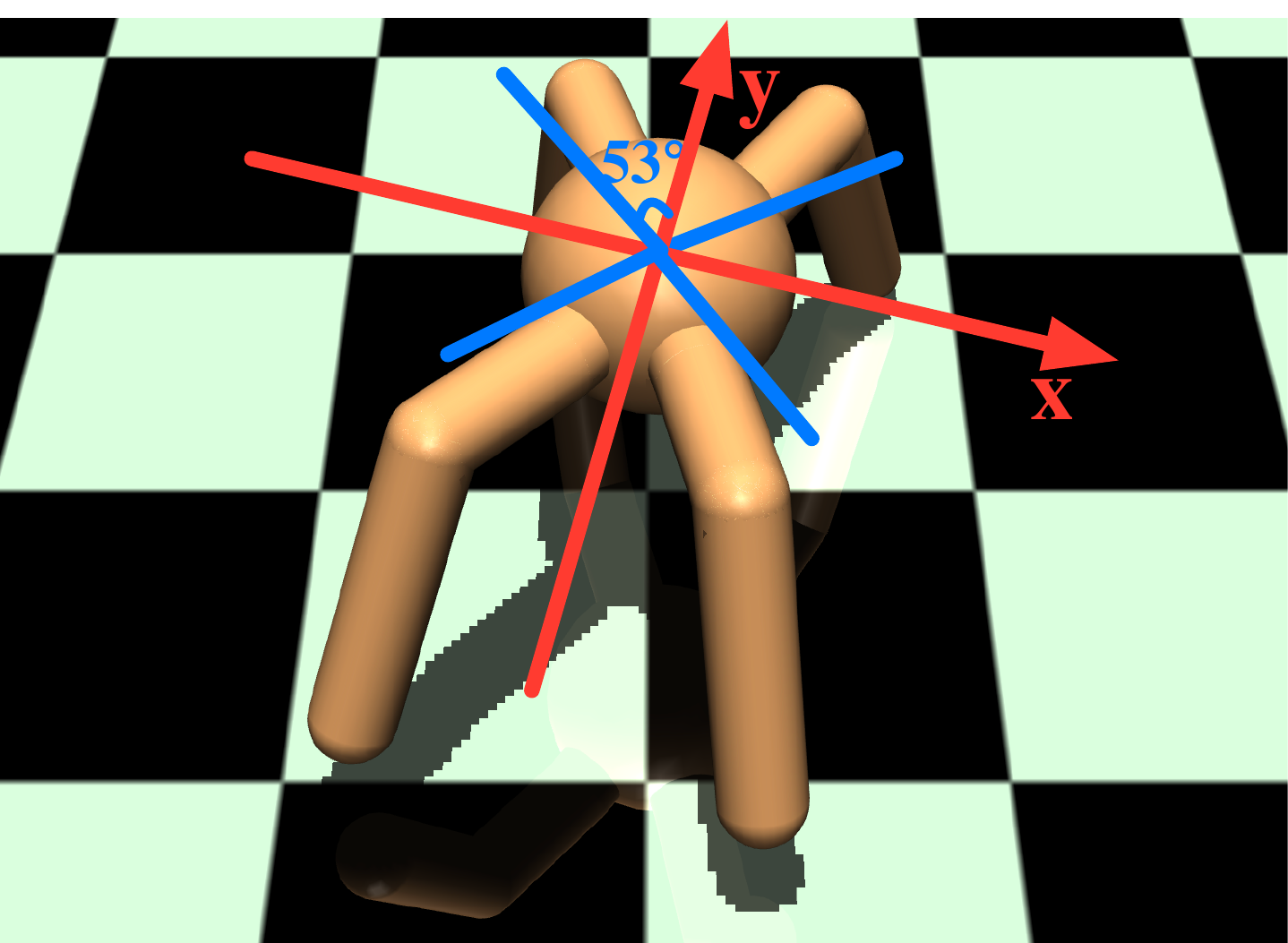}\label{fig:ant_y_env}}
    \subfigure[Swimmer BackDisabled]{\includegraphics[width=0.23\textwidth]{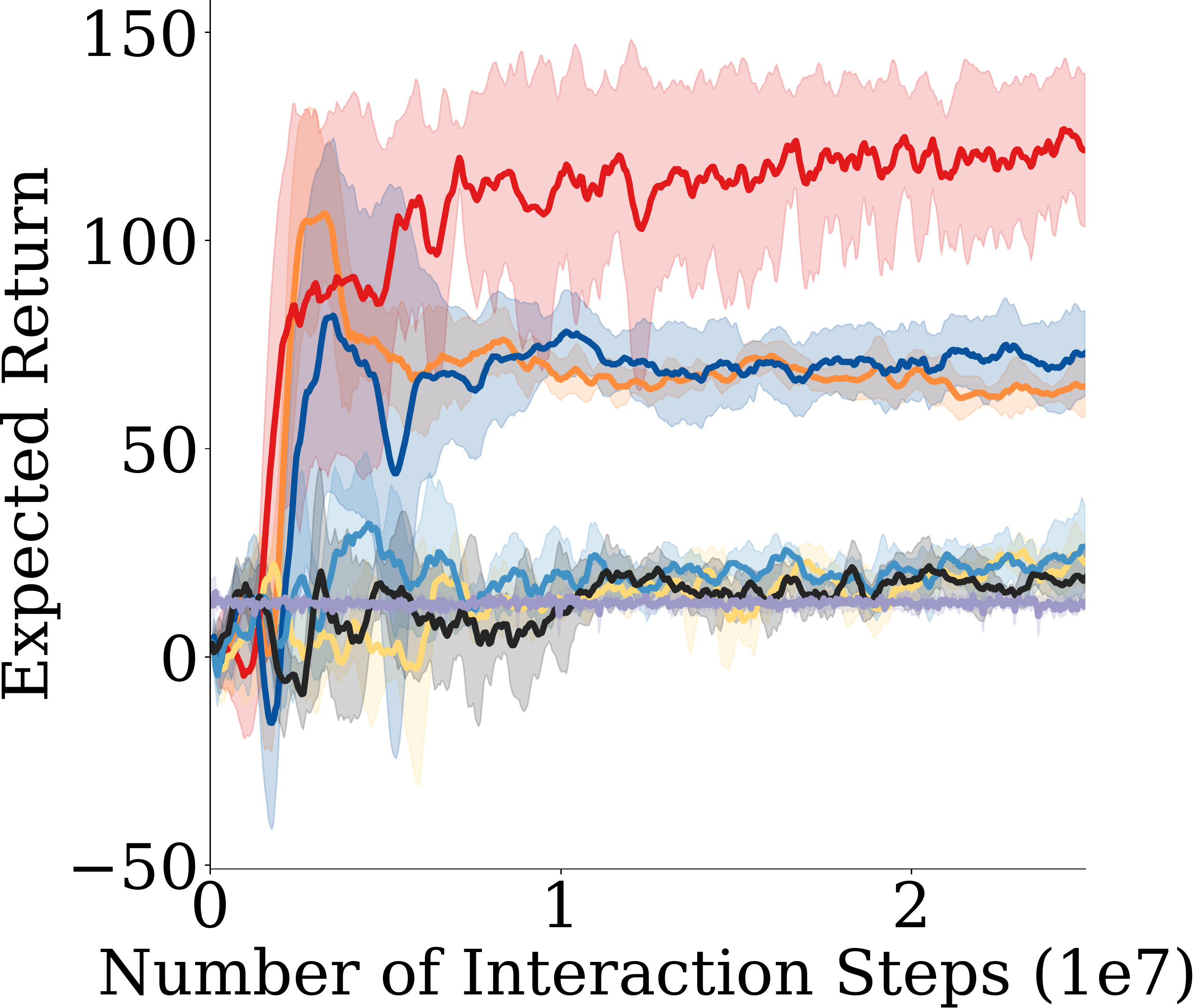}}
    \subfigure[Swimmer FrontDisabled]{\includegraphics[width=0.23\textwidth]{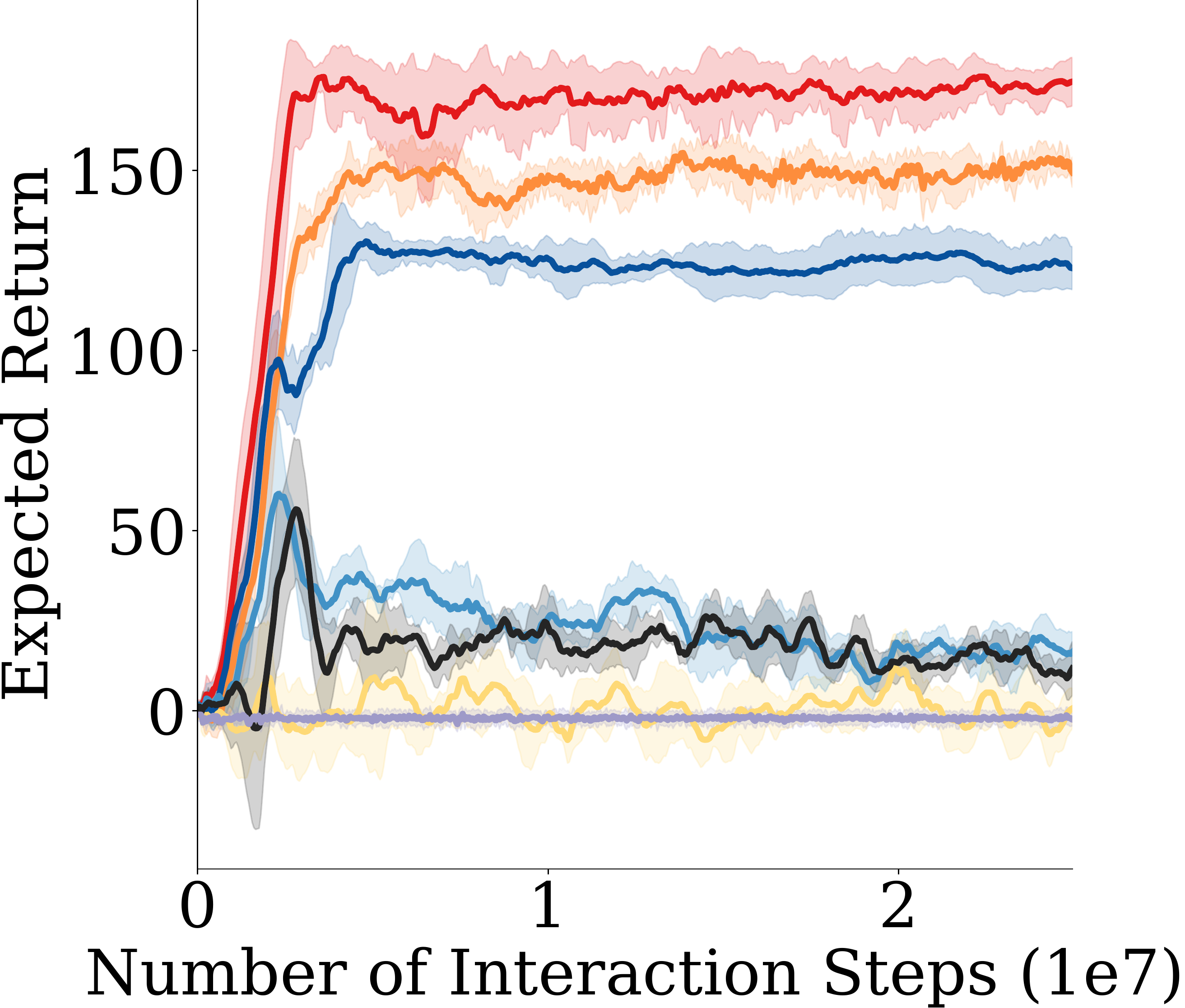}}
    \subfigure[Ant X-axis]{\includegraphics[width=0.23\textwidth]{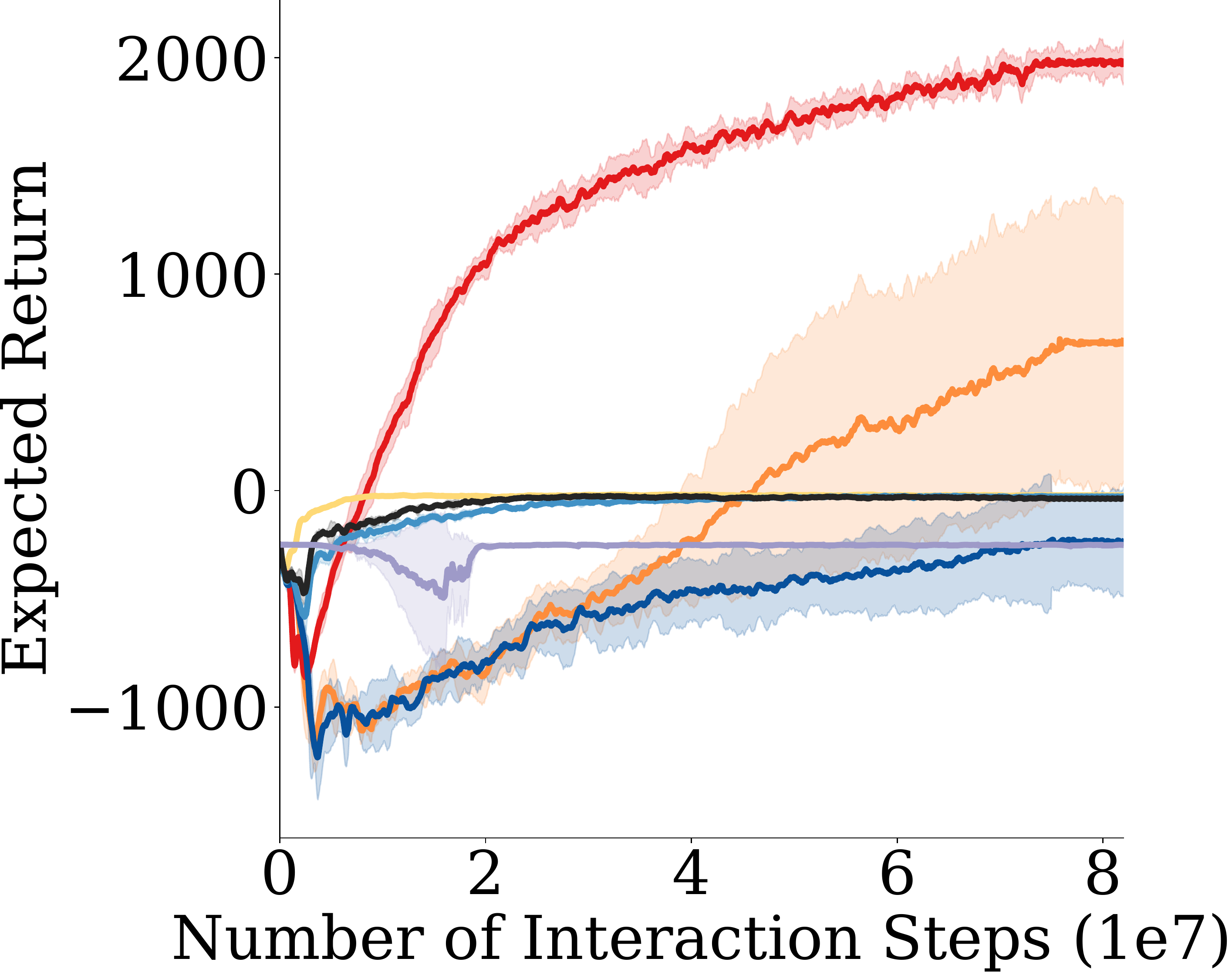}}
    \subfigure[Ant Y-axis]{\includegraphics[width=0.23\textwidth]{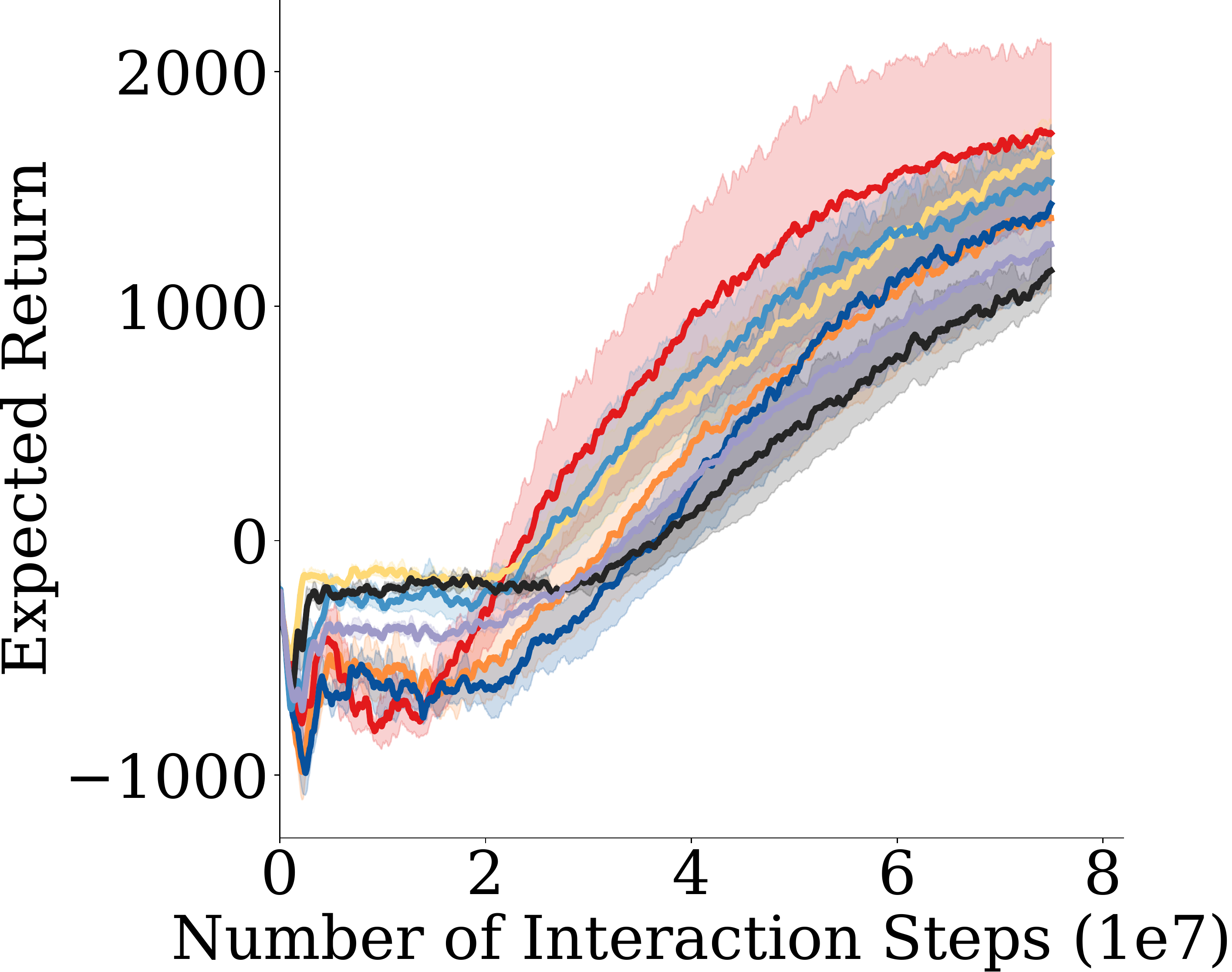}}
    \subfigure{\includegraphics[width=0.9\textwidth]{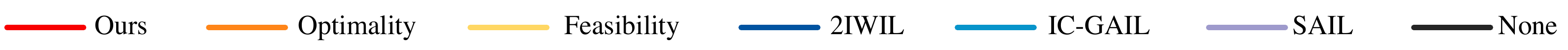}}
    \vspace{-12pt}
    \caption{(a-b) are the illustrations of Swimmer with disabled back joint and disabled front joint. (c-d) are the illustrations of Ant environment with  dynamics x-axis aligned and y-axis aligned dynamics. (e,f,g,h) are the expected return results on environments (a,b,c,d) respectively. One x-axis unit is $10^7$ steps.}
    \label{fig:result_swimmer_ant}
    \vspace{-15pt}
\end{figure*}

\noindent \textbf{Results.}
Fig.~\ref{fig:result_reacher_driving} shows that our proposed approach outperforms 2IWIL, IC-GAIL, and SAIL. 2IWIL and IC-GAIL focus on suboptimal demonstrations while SAIL focuses on demonstrations with varying dynamics. When both kinds of demonstrations are present, 2IWIL, IC-GAIL, and SAIL cannot learn a high return policy from the demonstrations. Our method employs feasibility and optimality to filter harmful demonstrations and enable successful imitation learning.

We also perform ablation studies on our method. In the driving and Reacher environments, since the expected return of the trajectory is influenced by the initial state, the proposed $f_{\text{rec}}$ is needed. So the optimality depends on the feasibility as shown in Eqn.~\eqref{eqn:frec}, and we cannot remove feasibility while preserving optimality.
We demonstrate the necessity of $f_{\text{rec}}$ by removing $f_{\text{rec}}$ and using a na\"ive optimality purely based on the reward (Na\"ive).
We observe that our method outperforms Na\"ive with a large margin in both environments. 
We then remove the optimality and only use the feasibility score. 
Our method outperforms Feasibility in both Reacher and Driving environments, indicating the importance of optimality to filter sub-optimal demonstrations. Note that Na\"ive performs similar to Feasibility. 
This is because in the environments where the expected return is influenced by the initial state, directly using expected return as the optimality tends to select initial states inducing high expected returns but ignore other useful initial states.
This can influence the diversity of the demonstrations and make the policy fail on low expected return initial states. 

\subsection{Environments with Similar Return across Initial States}
\noindent \textbf{Swimmer.} In the Swimmer environment, we have an agent with three links and two joints. 
The goal for the swimmer is to move forward by rotating the joints. As shown in Fig.~\ref{fig:swimmer_back_env} and Fig.~\ref{fig:swimmer_front_env}, we create two different dynamics: (1) the blue joint is disabled (BackDisabled) and (2) the red joint is disabled (FrontDisabled), where disabled means the joint angle is fixed to $0$ (the two links are in a line). 

\noindent \textbf{Ant.} In the Ant environment, we have an ant with four legs where each leg consists of two links and two joints. The goal for the ant is to move forward in the x-axis direction as fast as possible. As shown in Fig.~\ref{fig:ant_x_env} and Fig.~\ref{fig:ant_y_env}, we create different dynamics by setting the legs at different angles relative to each other: (1) the four legs can only rotate within $[0^\circ, 53^\circ]$ from the x-axis. and (2) the four legs can only rotate within $[0^\circ, 53^\circ]$ from the y-axis.

\noindent \textbf{Composition of Demonstrations.} The Swimmer and the Ant environments are two other interesting environments that we study. 
The ratio of demonstrations from optimal policy of the target environment, the sub-optimal policy of the target environment, and the optimal policy of a different environment is $1\%,49.5\%,49.5\%$ respectively for the Swimmer environment and for the Ant environment.

\noindent \textbf{Results.} 
In Fig.~\ref{fig:result_swimmer_ant}, we observe that the proposed method outperforms all the other methods with large margin in the expected return while having similar converging speed even on these complex tasks. 
2IWIL significantly outperforms other baseline methods.
This is because the influence of the sub-optimal demonstrations are larger than demonstrations from different dynamics.
2IWIL tends to filter more sub-optimal demonstrations, and thus performs better. However, 2IWIL still performs worse than our proposed method since our method further filters infeasible demonstrations, and learns from more feasible demonstrations from different dynamics.

The expected return of the trajectory is not influenced by the initial state for the Swimmer and the Ant environments, so the optimality does not depend on the feasibility. Therefore, our proposed optimality equals to using the raw expected return as the optimality score. Here, we run an ablation study, where we remove optimality and feasibility individually, denoted as Optimality and Feasibility.
We observe that our approach outperforms both Feasibility and Optimality, which demonstrates that both feasibility and optimality are important elements in selecting useful demonstrations.

\subsection{Robot Experiments}
\begin{figure}
    \centering
    \subfigure[Simulated Robot Arm]{\includegraphics[width=.23\textwidth]{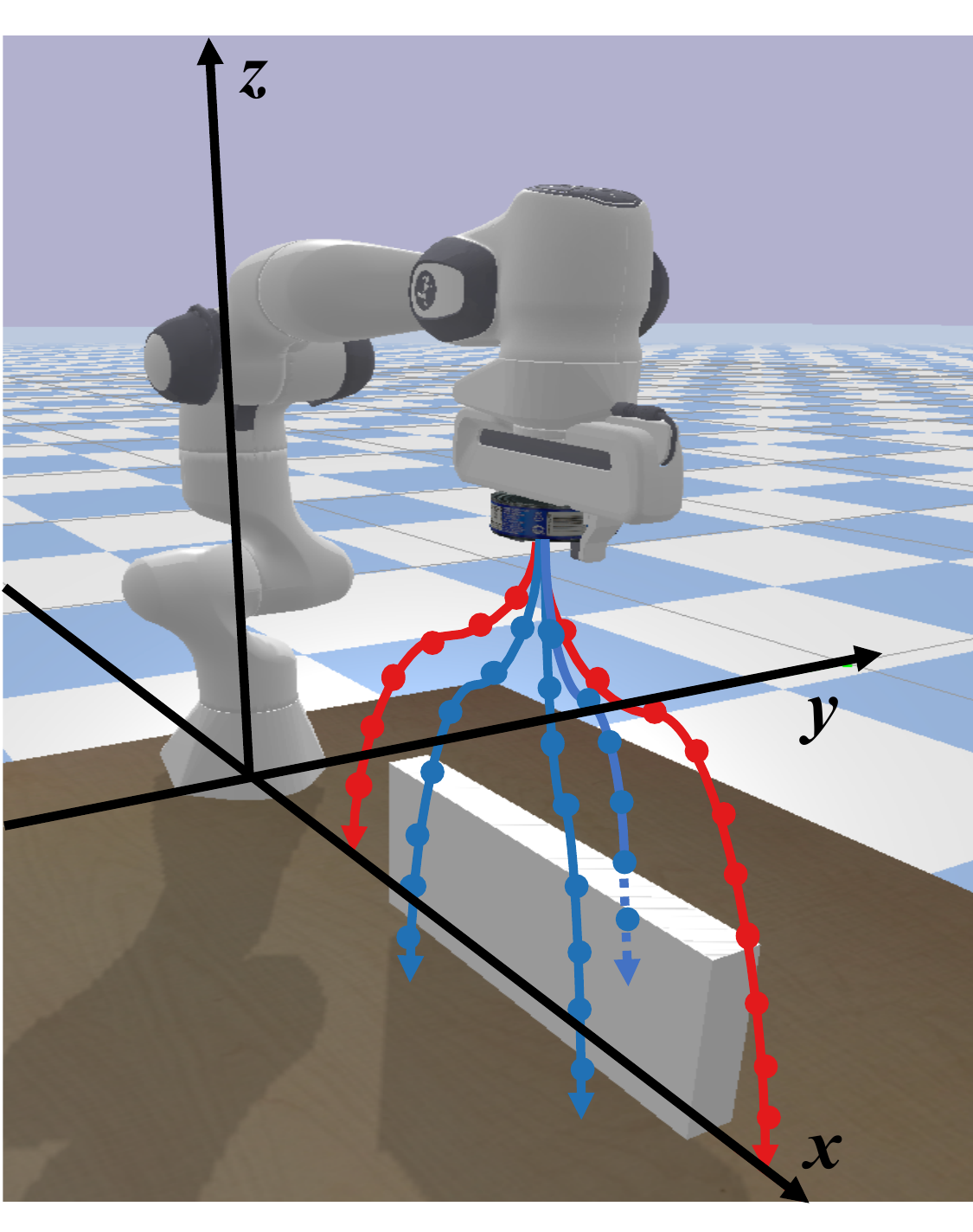}\label{fig:fake_robot}}
    \subfigure[Real Robot Arm]{\includegraphics[width=.23\textwidth]{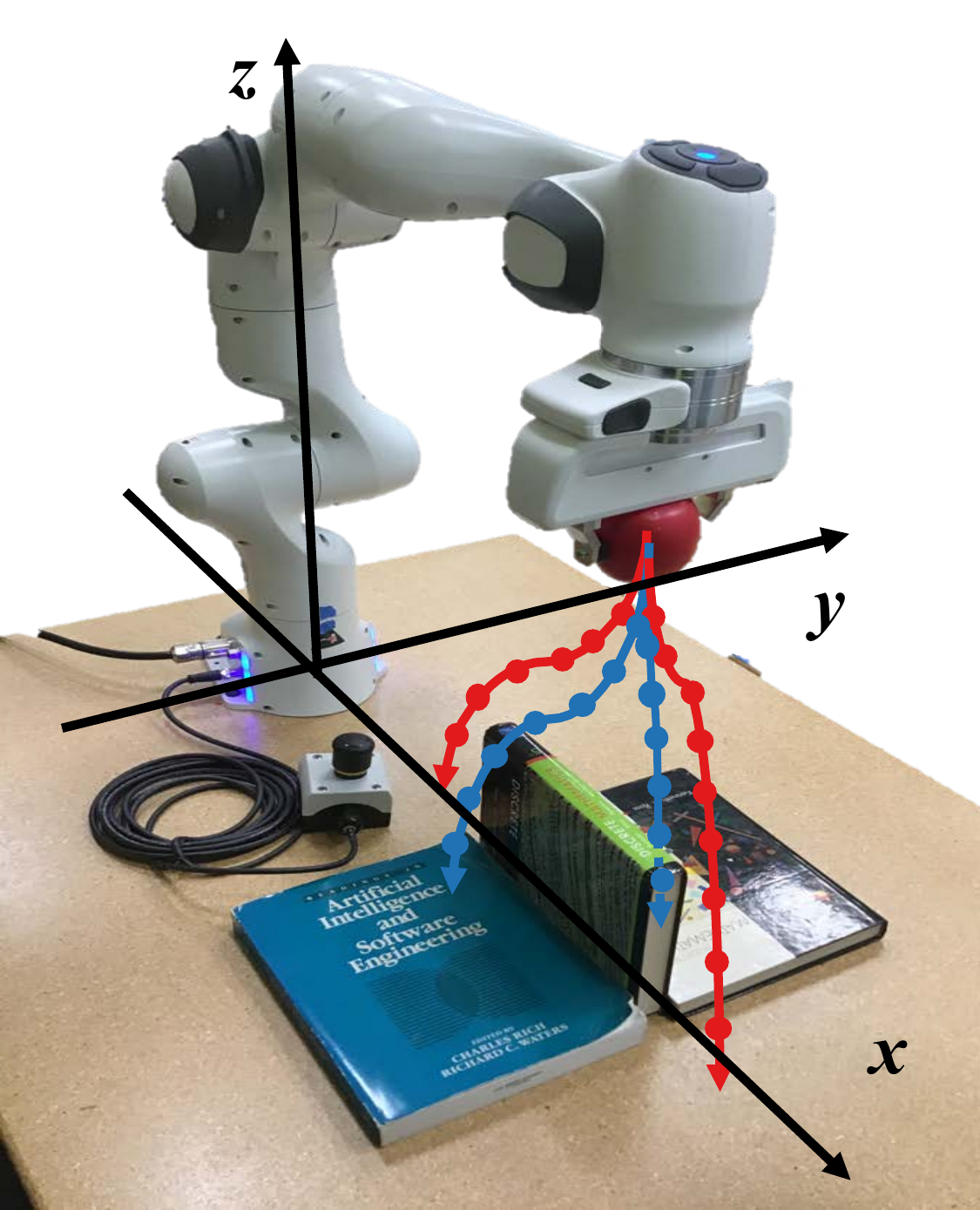}\label{fig:real_robot}}
    \subfigure[Simulated Robot Arm]{\includegraphics[width=.23\textwidth]{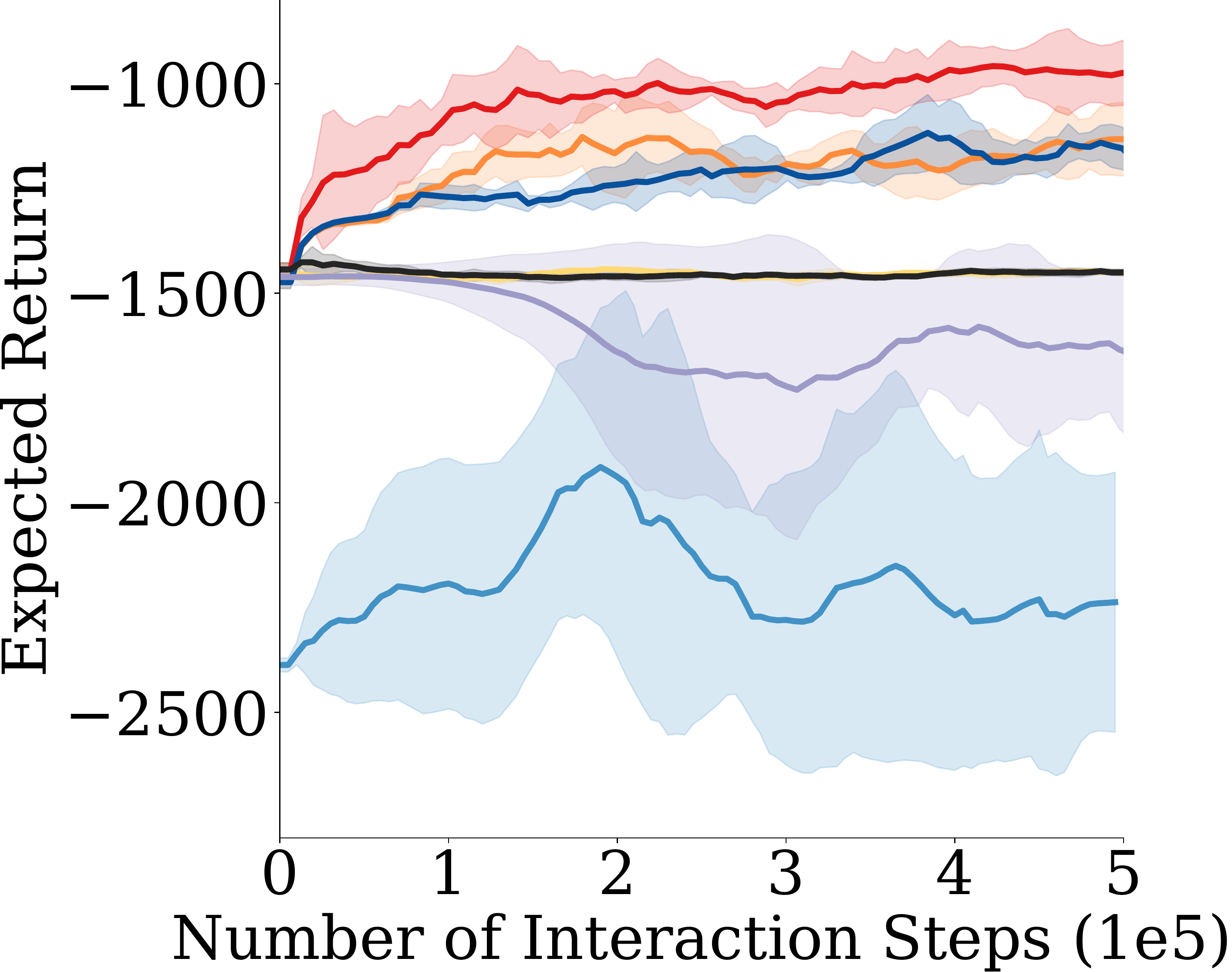}\label{fig:fake_robot_result}}
    \subfigure[Real Robot Arm]{\includegraphics[width=.23\textwidth]{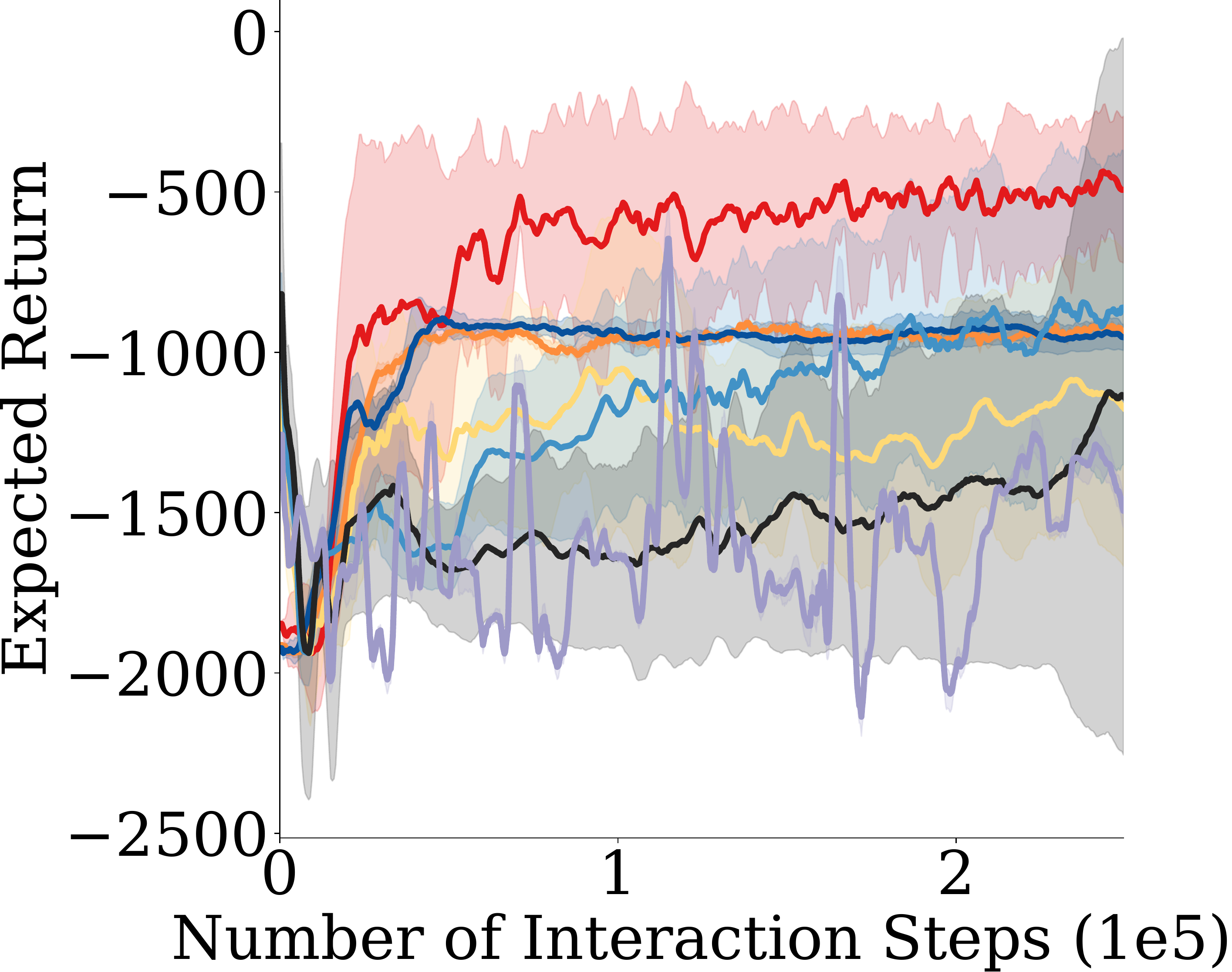}\label{fig:real_robot_result}}
    \subfigure{\includegraphics[width=0.44\textwidth]{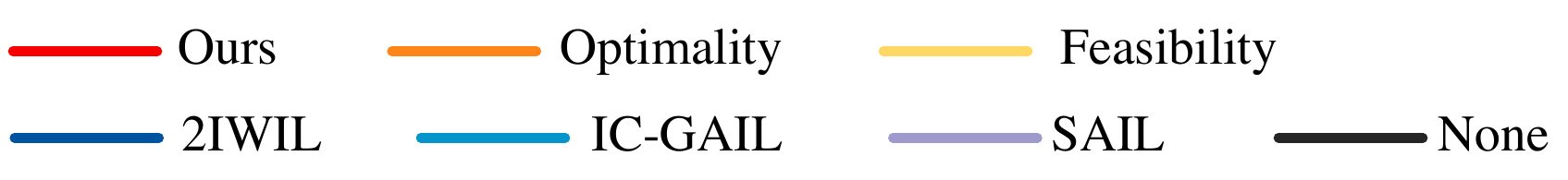}}
    \vspace{-10pt}
    \caption{(a-b) are the illustrations of a simulated robot arm and a real robot arm, where the blue curves are demonstrated by the 7-DoF arm, while the 3-DoF arm can only demonstrate the red curves. (c-d) are the corresponding expected returns. One x-axis unit is $10^5$ steps.}
    \label{fig:result_robot_arm}
    \vspace{-15pt}
\end{figure}

The Panda Robot Arm environment uses a $7$ Degrees of Freedom (DoF) Franka Panda Robot arm~\cite{frankapandarobot}. 
We create two dynamics: (1) the original 7-DoF arm and (2) a 3-DoF arm by fixing all the joints that rotate around the z-axis. 
As shown in Fig.~\ref{fig:fake_robot} and Fig.~\ref{fig:real_robot}, we conduct experiments on a task, where we want the robot arm to take an object from a starting configuration above the table and move to the table without colliding with the obstacle. The 7-DoF arm can have various paths shown in blue trajectories while the 3-DoF robot can only move along the x-axis resulting in either one of the two red trajectories to avoid the obstacle. So different dynamics can introduce different optimal trajectories. 

\noindent \textbf{Composition of Demonstrations.} The ratio of demonstrations from optimal policy of the target environment, the suboptimal policy of the target environment, and the optimal policy of a different environment is $1\%,49.5\%,49.5\%$. 

\noindent \textbf{Results.}
In the robot environment the initial state does not influence the highest expected return much, so we compare the proposed method with Feasibility and Optimality similar to the Swimmer and Ant environments.
We conduct experiments both on a simulated panda robot arm with pybullet~\cite{coumans2018pybullet} and a real robot arm. 
In Fig.~\ref{fig:result_robot_arm}, we observe that the proposed method outperforms all the baselines with a large margin. 
Our method outperforms the variants with only optimality or only feasibility considered individually, which demonstrates the necessity of both feasibility and optimality in realistic environments.

\section{Conclusion}
\noindent \textbf{Summary.} In this paper, we propose an algorithm to learn from imperfect demonstrations from agents with different dynamics, where the demonstrations can be drawn from non-expert policies or from agents with different dynamics. 
We design a feasibility score and an optimality score to select demonstrations that are useful for the target agent.
We show that the policy learned from the selected demonstrations outperforms other imitation learning methods on various environments and different composition of demonstrations.

\noindent \textbf{Limitations and Future Work.} 
Although our work relaxes stringent assumptions placed on imitation learning algorithms, it is also limited in a few ways:
To compute feasibility, our method uses the inverse dynamics function and the environment for trajectory sampling, which might not be accessible in some settings. 
To compute the rectify function $f_{\text{rec}}$, we need the neighborhood property, which may also be violated in particular environments.
For example, in navigation, if two nearby locations are separated by an obstacle, their navigation route can differ significantly and the highest expected return might also be quite different. 
In the future, we plan to address these challenges by incorporating structures about the problem that can help us better learn the inverse dynamics or rectify functions that do not rely on the neighborhood property.

\section{Acknowledgements}
We would like to thank FLI grant RFP2-000 and NSF Award Number 1849952 for their support.

{\small
\bibliography{main.bib}
\bibliographystyle{IEEEtran}
}

\appendix

\section{Algorithm}
In Algorithm~\ref{alg:algo}, we go through the steps of the algorithm for learning from imperfect demonstrations from agents with varying dynamics. 

\begin{algorithm}[h]
\KwIn{Demonstrations $\Xi$, random trajectories from the target agent $\Xi_f$}
 Compute the feasibility $w_f$ $\forall \xi$ in $\Xi$ using Eqn. (3) \\
 Compute the optimality $\forall \xi$ in $\Xi$ using Eqn. (5)\\
 Compute the state transition distribution $p_w$ with the probability proportion to the score $w$ in Eqn. (7) \\
 \While{not converging}{
     Train $\pi$ with an state-based imitation learning algorithm with state transitions sampled from $p_w$\;
  }
  \KwOut{Learned optimal policy $\pi^*$ for $\mathcal{M}$.}
  \caption{Algorithm}\label{alg:algo}
\end{algorithm}

\section{Setting Threshold}
We set two threshold values for the distance between trajectories. First, $d_{\text{min}}$ sets the minimum distance between $\xi$ and $\xi'$, and any trajectory with distance smaller than $d_{\text{min}}$ will be assigned feasibility of $1$.
We set $d_{\text{min}}$ by taking the minimum in the sampled feasible trajectories $\Xi_f$: $d_{\text{min}} = \min_{\xi_f\in \Xi_f } F(\xi_f, \xi'_f)$. 

Second, $d_{\text{max}}$ sets the maximum distance between $\xi$ and $\xi'$ that can be tolerated.
There are two clear approaches for computing $d_{\text{max}}$: the maximum distance in the demonstration trajectories $\max_{\xi \in \Xi} F(\xi, \xi')$ and the maximum distance in the sampled feasible trajectories $\max_{\xi_f \in \Xi_f} F(\xi_f, \xi'_f)$. 
Here, $\xi'$ and $\xi'_f$ are computed using the inverse dynamics as in Eqn.~(1) in the main text.
 The former can be too large and can assign positive feasibility to harmful and completely infeasible trajectories at times.
 On the other hand, the latter may filter nearly feasible trajectories, which are useful for imitation learning. 
Instead of these two extreme measures, we propose estimating $d_{\text{max}}$ from the environment. 
 For every trajectory $\xi_f$ in $\Xi_f$, we derive a trajectory $\xi''_f=\{s''_{0},s''_1,\cdots, s''_N\}$:

\begin{equation}\label{eqn:delta_s}
\begin{aligned}
    & s''_0 = s_0, \quad a''_t = f_{\text{id}}(s''_{t-1}, s_{t}) , t\ge1 \\
    & s^{\text{sample}}_{t} \sim p(s''_{t-1}, a''_t, s''_{t}), t\ge1\\
    & s''_{t} = s^{\text{sample}}_{t} + \Delta s.
\end{aligned}
\end{equation}
We perturb the state at each step with a $\Delta s$, where $\lvert\Delta s \rvert \le \delta$, and then we set $d_{\text{max}}$ as the max distance $\max_{\xi_f \in \Xi_f}F(\xi_f, \xi''_f)$. We note that $\xi'_f$ is the corresponding trajectory for $\xi$ while $\xi''_f$ is a trajectory with perturbation $\delta s$ over the states of $\xi$.
This modeling allows us to know that a distance smaller than $d_{\text{max}}$ means the trajectory is within $\delta$ state averaged distance from a feasible trajectory. 
We can control $\delta$ in different environments to decide if a trajectory should be assigned positive or zero feasibility weights.

\begin{table*}[htb]
    \centering
    \caption{Expected return after convergence for all the environments.}
    \label{tab:table_results}
    \resizebox{\textwidth}{!}{
    \begin{tabular}{c|c c|c c|c c|c c|c c}
    \hline
    \noalign{\vskip 2pt}    
    \multirow{3}{40pt}{Methods} & \multicolumn{10}{c}{Environments} \\
    \noalign{\vskip 2pt}    
    \cline{2-11}
    \noalign{\vskip 2pt}    

         & \multicolumn{2}{c|}{Reacher} &  \multicolumn{2}{c|}{Driving} & \multicolumn{2}{c|}{Swimmer} &  \multicolumn{2}{c|}{Ant} & \multicolumn{2}{c}{Panda}\\
        \noalign{\vskip 2pt}   
         \cline{2-11}
        \noalign{\vskip 2pt}
         & Clockwise & Counter Clockwise & Slow & Fast & BackDisabled & FrontDisabled & X-axis & Y-axis & Simulation & Real \\
             \noalign{\vskip 2pt}
         \hline
             \noalign{\vskip 2pt}    
         None & -122.51$\pm$5.98 & -127.68$\pm$9.46 & -6218$\pm$179 & -6232$\pm$69 & 19.57$\pm$6.98 & 11.89$\pm$2.64 & -48$\pm$8 & 1152$\pm$85 & -1415$\pm$9 & -1177$\pm$1156\\
         SAIL & -129.52$\pm$6.83 & -129.43$\pm$8.12 & -5743$\pm$1043 & -8976$\pm$1772 & 12.51$\pm$0.43 & -2.63$\pm$1.11 & -255$\pm$5 & 1239$\pm$92 & -1645$\pm$208 & -1498$\pm$78\\
         IC-GAIL & -121.48$\pm$14.56 & -126.95$\pm$10.26 & -8189$\pm$1696 & -7202$\pm$403 & 25.69$\pm$8.39 & 14.56$\pm$5.79 & -46$\pm$6 & 1532$\pm$133 & -2241$\pm$273 & -876$\pm$503\\
         2IWIL & -134.68$\pm$2.78 & -118.92$\pm$5.54 & -5189$\pm$1754 & -5632$\pm$84 & 74.12$\pm$9.14 & 123.89$\pm$5.12 & -247$\pm$241 & 1449$\pm$297 & -1148$\pm$22 & -915$\pm$96\\
         Feasibility & -122.03$\pm$5.46 & -121.32$\pm$7.53 & -7364$\pm$455 & -7188$\pm$845 & 23.89$\pm$4.77 & 0.32$\pm$6.32 & -42$\pm$10 & 1621$\pm$94 & -1412$\pm$6 & -1203$\pm$412\\
         Na\"ive & -111.53$\pm$3.85 & -118.96$\pm$10.89 & -4812$\pm$765 & -5913$\pm$1924 & - & - & - & - & - & -\\
         Optimality & - & - & - & - & 63.12$\pm$5.01 & 150.32$\pm$3.45 & 689$\pm$648 & 1389$\pm$356 & -1124$\pm$85 & -902$\pm$23\\
         \hline
         Ours & \textbf{-94.32}$\pm$9.67 & \textbf{-90.89}$\pm$4.02 & \textbf{-654}$\pm$103 & \textbf{-1429}$\pm$672 & \textbf{122.44}$\pm$13.42 & \textbf{167.43}$\pm$6.23 & \textbf{1979}$\pm$83 & \textbf{1709}$\pm$412 & \textbf{-979}$\pm$89 & \textbf{-500}$\pm$201\\
         \hline
    \end{tabular}
    }
    
\end{table*}

\section{Time Complexity}
Compared to vanilla imitation learning, our approach introduces additional computation costs in computing $w$ and $p_w$. The computational cost of vanilla imitation learning is bounded as shown in~\cite{ho2016generative,torabi2018behavioral,torabi2018generative}. So here we only need to demonstrate that computing $w$ and $p_w$ is tractable.

We let the size of the demonstration set $\Xi$ be $D$, the size of the feasible trajectory set $\Xi_f$ be $D_f$, and the largest number of steps of a trajectory in $\Xi$ and $\Xi_f$ be $N$. We let $\text{dim}(s)$ be the dimension of the state. The average time to compute the inverse dynamics function is $T_{\text{id}}$. 

For the feasibility, we first derive the feasible trajectory set $\Xi_f$, and the time complexity is $O(D_fN)$. Then we compute the corresponding trajectory $\xi'$ for each trajectory $\xi$ in $\Xi$ and the time complexity is $O(DNT_{\text{id}})$. With the corresponding trajectory, we then compute $F(\xi,\xi')$ as shown in Equation (3) in the main text for all trajectories in $\Xi$, and the time complexity is $O(DN\text{dim}(s))$ for iterating over all the states in the trajectory. Then we assess the time to compute the two thresholds $d_{\text{max}}$ and $d_{\text{min}}$. $d_{\text{max}}$ needs simulating the trajectories in $\Xi_f$ with perturbation $\delta s$ and computing $F(\xi_f,\xi''_f)$, so the time to compute Equation~\eqref{eqn:delta_s} is $O(D_fN(T_{\text{id}}+\text{dim}(s)))$. $d_{\text{min}}$ needs simulating $\xi'_f$ with each trajectory $\xi_f\in\Xi_f$ and computing $F(\xi_f,\xi'_f)$ as shown in Equation (1) of the main text, so the time is $O(D_fN(T_{\text{id}}+\text{dim}(s)))$. Therefore, the total complexity time is the sum of the above computations:  $O((D_f+D)N(T_{\text{id}}+\text{dim}(s)))$.

For the optimality, to estimate the function $f_{\text{rec}}$, we need to compute the distance between the initial states of every pair of trajectories. So the time complexity is $O(D^2\text{dim}(s))$. The time to compute $w_o$ is $O(D\text{dim}(s))$. As a result the total time complexity is $O(D^2\text{dim}(s))$. 

Therefore, the total time to compute the proposed $w$ is $O(D^2\text{dim}(s)+(D_f+D)N(T_{\text{id}}+\text{dim}(s)))$. The time complexity increases linearly with respect to the dimension of the state, which depends on the degrees of freedom of the agent. 

In practice, the time to compute the inverse dynamics is negligible. We would like to emphasize that most imitation learning algorithms traverse over the demonstrations multiple times, and the time complexity of traversing over trajectories once is $O(DN\text{dim}(s))$.
Therefore, compared to the time for running imitation learning algorithms, the time cost $O(DN(T_{\text{id}}+\text{dim}(s)))$ is negligible. 

So the main time burden is on $O(D^2\text{dim}(s))$ and $O(D_fN(T_{\text{id}}+\text{dim}(s)))$. In practice, if the size of the demonstration set $D$ is too large, when estimating $f_{\text{rec}}$, we can sample a subset of demonstrations to reduce the time complexity while still obtaining an relatively accurate estimation of $f_{\text{rec}}$. $D_fN$ means how many feasible trajectories we collect in $\Xi_f$, which is decided by the environment. More complex environments need more feasible trajectories to cover the trajectory space. To reduce $D_fN$, we can first learn an initial policy by direct imitation learning from all the demonstrations and use that policy to collect feasible trajectories. These trajectories are likely to be more close to the optimal trajectories and we can avoid wasting time to explore the trajectories that are far from optimal trajectories.

\section{Implementation Details}
For the network architecture, we use a three-layer fully-connected network with hidden state size $100$ and tanh activation function for the policy network, the value network and the discriminator (the value network and the discriminator are only needed for GAIL-based methods). 
For the inverse dynamics model, we use an eight-layer fully-connected network with ReLU activation. For the behavior-cloning-based methods, we minimize the smooth l1 loss between the predicted action and the ground truth action. 
For the GAIL-based methods, we employ TRPO-based GAIL. We use an Adam optimizer~\cite{kingma2014adam}, fix the learning rate for the policy network and the discriminator as $10^{-3}$, and fix the value network as $3\times 10^{-4}$ for all the environments. For the two hyper-parameters $\Delta s$ and $\sigma$, we tune the parameters for one task in each environment while fixed the parameters for the other tasks in the environment. We set $\Delta s=0.001$ for $\sigma=50$ for Reacher, $\Delta s=0.001$ for $\sigma=100$ for Driving, $\Delta s=0.0005$ for $\sigma=40$ for Swimmer, and $\Delta s=0.001$ for $\sigma=500$ for Ant and We set $\Delta s=0.1$ for $\sigma=100$ for Robot Arm. 

\section{Tabularised results}
To more clearly compare the expected return of different methods, we show the expected return for each methods and environments after convergence in Table~\ref{tab:table_results}. We can observe that the proposed approach achieves the highest expected return after convergence. 

\begin{figure}
    \centering
    \includegraphics[width=.45\textwidth]{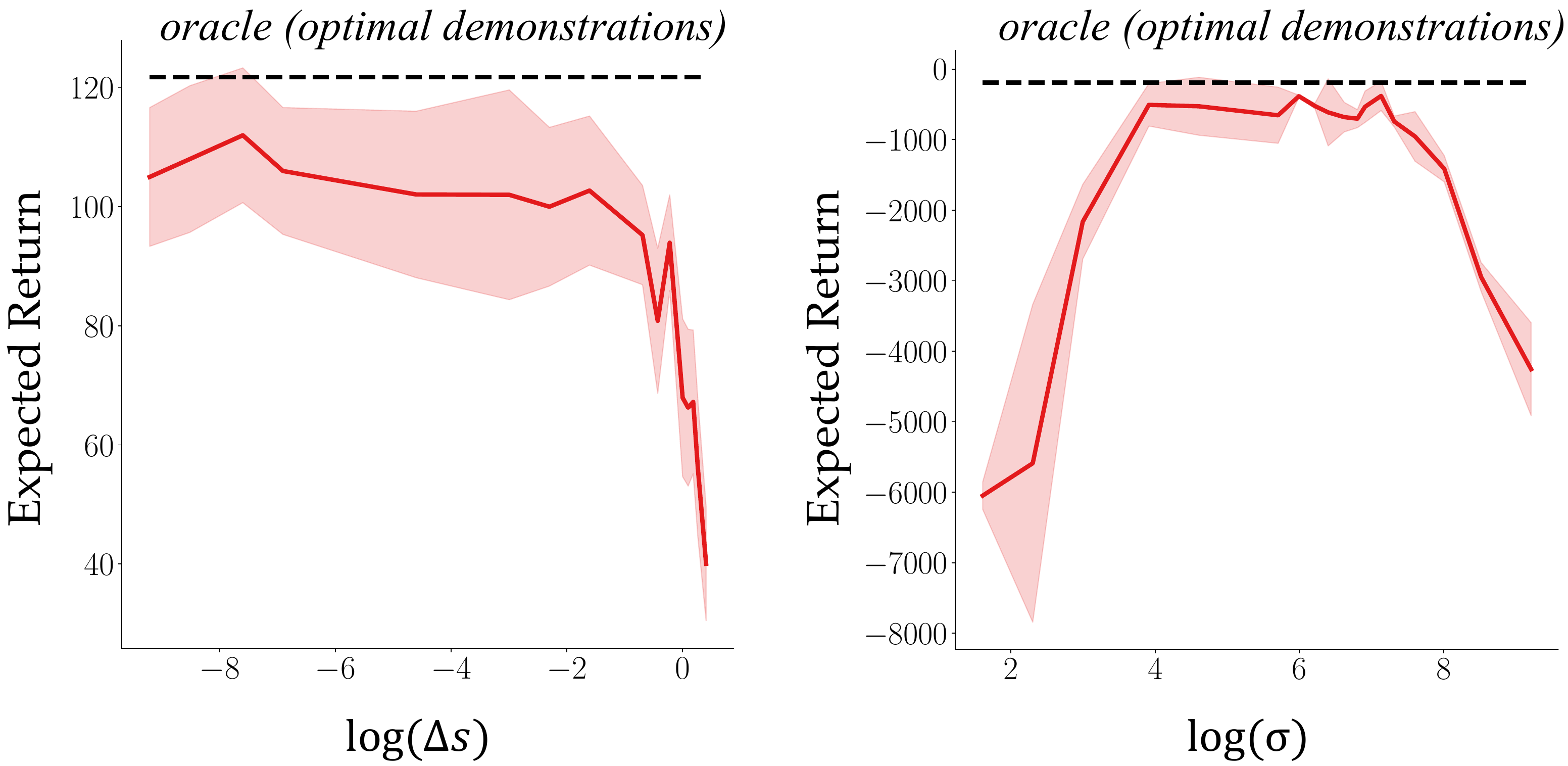}
    \caption{(left) Expected return with respect to different $\log(\Delta s)$ test on the BackDisabled Swimmer environment; (right) Expected return with respect to different $\log(\sigma)$ test on the Slow Driving environment.}
    \label{fig:sensitivity}
\end{figure}

\section{Parameter Sensitivity}
We have two key hyper-parameters to set in our method: the $\Delta s$ in Eqn.~(4) and the $\sigma$ in Eqn.~(5).
We note that the two hyper-parameters are necessary to achieve the optimal performance of the method but the algorithm can work stably in a reasonable range for these hyper-parameters. We run the experiments on the BackDisabled Swimmer environment with respect to different $\log(\Delta s)$ and on the Slow Driving environment with respect to different $\log(\sigma)$. We use these two environments because Slow Driving uses our full version of optimality with the rectify function while Swimmer relies more on the feasibility. The solid red line is the performance of the proposed method while the dashed black line is the expected return of the optimal demonstrations.
Fig.~\ref{fig:sensitivity} shows that there exist a wide range of values for both $\Delta s$ and $\sigma$ to lead to stable results and high expected return. 
However, when $\Delta s$ is too large, $d_{\text{max}}$ can become larger and lead to positive feasibility be assigned to infeasible demonstrations. On a similar note, if $\sigma$ is too large or too small, the performance drops severely. 
Small values of $\sigma$ can cause a large difference in optimality for similar rectified rewards. This can overweigh some optimal demonstrations over others, while too large values of $\sigma$ can lead to the optimality failing to discriminate between optimal demonstrations and sub-optimal ones.

\begin{figure}
    \centering
    \subfigure[Vary Perfect Demonstrations]{\includegraphics[width=0.23\textwidth]{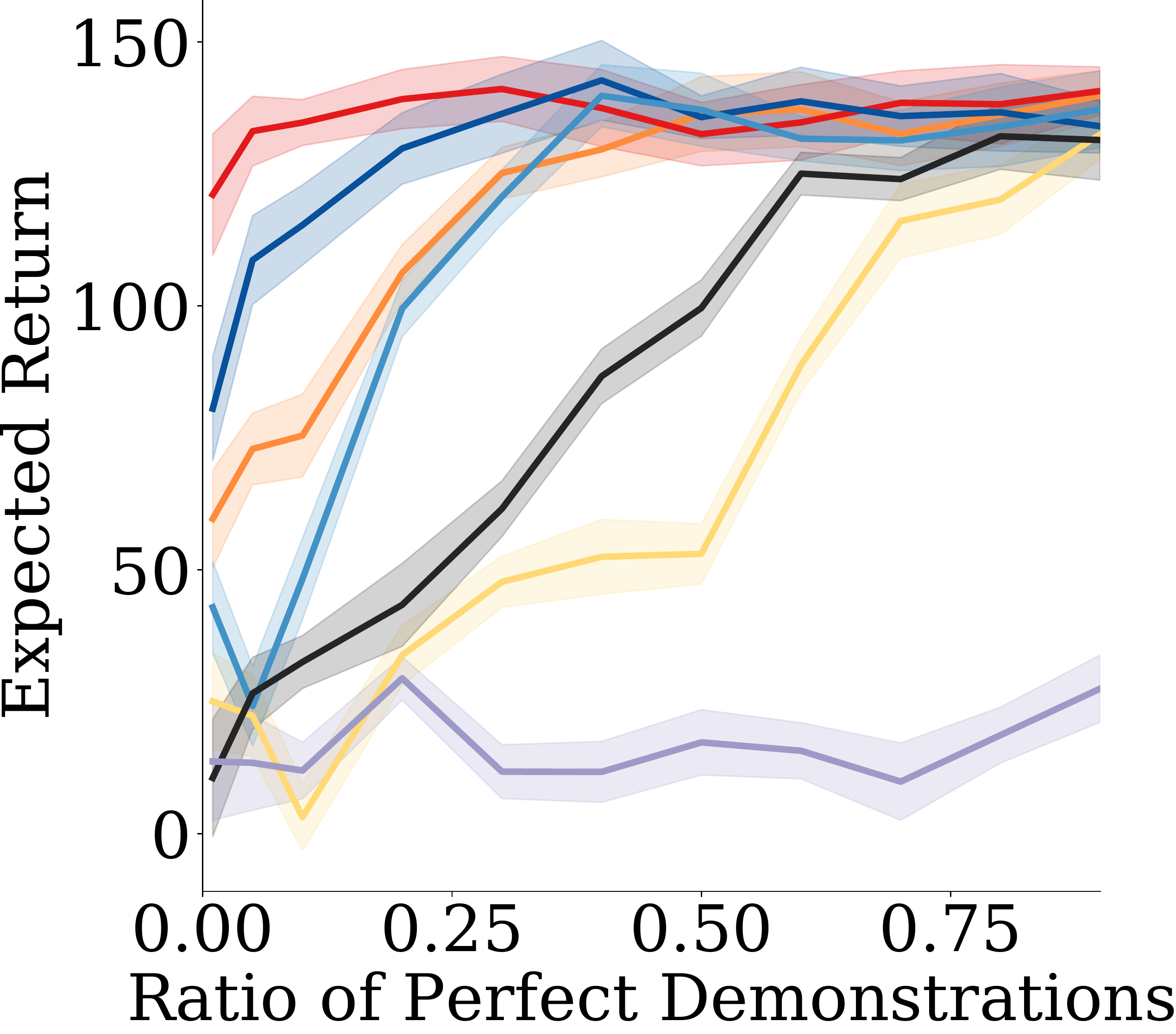}\label{fig:varying_perfect_portion}}
    \subfigure[Vary Imperfect Demonstrations]{\includegraphics[width=0.23\textwidth]{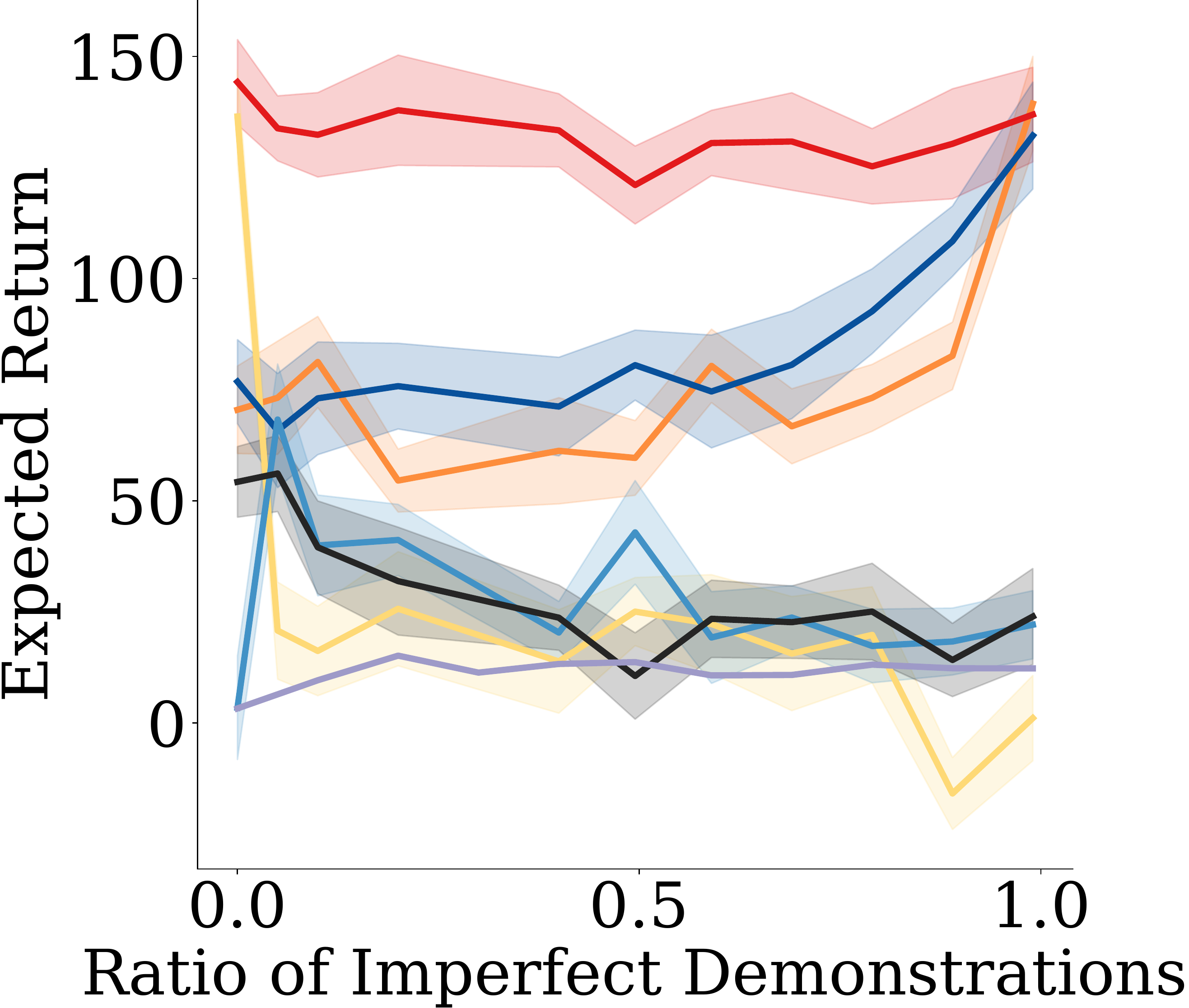}\label{fig:varying_imperfect_portion}}
    \subfigure{\includegraphics[width=0.46\textwidth]{figs/legend_conf1.pdf}}
    \caption{Ratio of Perfect, Infeasible, and Suboptimal Demonstrations. In (a), we change the ratio of perfect demonstrations, while keeping the ratio of infeasible demonstrations equal to the ratio of suboptimal demonstrations. In (b), we fix the ratio of perfect demonstrations, while changing the ratio of infeasible demonstrations and suboptimal demonstrations. 
    }
    \label{fig:varying_portion}
\end{figure}

\section{Varying Composition of Demonstrations}
\noindent \textbf{Effects of the Ratio of Perfect Demonstrations.} 
Here we conduct an experiment specifically on the Swimmer environment, to demonstrate the effects of gradually increasing the portion of perfect demonstrations -- demonstrations provided by an optimal policy on the target MDP -- while keeping the relative size of other types of demonstrations constant.
We observe that our method consistently performs well with different ratios of perfect demonstrations as opposed to all the other methods where performance significantly drops with small ratio of perfect demonstrations (see Fig.~\ref{fig:varying_perfect_portion}). 
When the ratio of perfect demonstrations becomes high enough, the suboptimal and infeasible demonstrations have no influence on learning and even vanilla GAIL learns the optimal policy.

\noindent \textbf{Effects of the Relative Ratio of Demonstrations with Varying Dynamics and Suboptimal Demonstrations.}
We conduct an experiment in the Swimmer environment, where we vary the ratio of demonstrations with varying dynamics and suboptimal demonstrations.
We fix the portion of perfect demonstrations at $1\%$ of all trajectories.
In Fig.~\ref{fig:varying_imperfect_portion}, we observe that our method consistently outperforms other methods for all ratios of these suboptimal demonstrations.
We also observe that the performance of the other methods including Optimality increases with increasing the ratio of imperfect demonstrations. However, the performance of Feasibility decreases, which demonstrates that our proposed optimality can handle under learning with sub-optimal demonstrations and our proposed feasibility score can learn from demonstrations from agents with varying dynamics. 

\end{document}